# Separating populations with wide data: A spectral analysis

**Avrim Blum**[*]

*Computer Sciene Departement
Carnegie Mellon University
5000 Forbes Ave
Pittsburgh, PA 15213, USA
e-mail:* avrim@cs.cmu.edu

**Amin Coja-Oghlan**[†]

*University of Edinburgh
Informatics Forum
10 Crichton Street
Edinburgh EH8 9AB, UK
e-mail:* acoghlan@inf.ed.ac.uk

**Alan Frieze**[‡]

*Departement of Mathmatics
Carnegie Mellon University
5000 Forbes Ave
Pittsburgh, PA 15213, USA
e-mail:* alan@random.math.cmu.edu

**Shuheng Zhou**[§]

*Seminar für Statistik
ETH Zentrum,
CH-8092 Zürich, Switzerland
e-mail:* zhou@stat.math.ethz.ch

**Abstract:** In this paper, we consider the problem of partitioning a small data sample drawn from a mixture of $k$ product distributions. We are interested in the case that individual features are of low average quality $\gamma$, and we want to use as few of them as possible to correctly partition the sample. We analyze a spectral technique that is able to approximately optimize the total data size—the product of number of data points $n$ and the number of features $K$—needed to correctly perform this partitioning as a function of $1/\gamma$ for $K > n$. Our goal is motivated by an application in clustering individuals according to their population of origin using markers, when the divergence between any two of the populations is small.

[*]Supported in part by the NSF under grant CCF-0514922
[†]Supported by the German Research Foundation under grant CO 646
[‡]Supported in part by the NSF under grant CCF-0502793
[§]Supported in part by the NSF under grants CCF-0625879 and CNF–0435382

**Contents**

**1. Introduction**

We explore a type of classification problem that arises in the context of computational biology. The problem is that we are given a small sample of size $n$, e.g., DNA of $n$ individuals (think of $n$ in the hundreds or thousands), each described by the values of $K$ *features* or *markers*, e.g., SNPs (Single Nucleotide Polymorphisms, think of $K$ as an order of magnitude larger than $n$). Our goal is to use these features to classify the individuals according to their population of origin. Features have slightly different probabilities depending on which population the individual belongs to, and are assumed to be independent of each other (i.e., our data is a small sample from a mixture of $k$ very similar product distributions). The objective we consider is to minimize the total data size $D = nK$ needed to correctly classify the individuals in the sample as a function of the "average quality" $\gamma$ of the features, under the assumption that $K > n$. Throughout the paper, we use $p_i^j$ and $\mu_i^j$ as shorthands for $p_i^{(j)}$ and $\mu_i^{(j)}$ respectively.

**Statistical Model:** We have $k$ probability spaces $\Omega_1, \ldots, \Omega_k$ over the set $\{0,1\}^K$. Further, the components (*features*) of $z \in \Omega_t$ are independent and $\mathbf{Pr}_{\Omega_t}[z_i = 1] = p_t^i$ ($1 \leq t \leq k$, $1 \leq i \leq K$). Hence, the probability spaces $\Omega_1, \ldots, \Omega_k$ comprise the distribution of the features for each of the $k$ populations. Moreover, the input of the algorithm consists of a collection (*mixture*) of $n = \sum_{t=1}^k N_t$ unlabeled samples, $N_t$ points from $\Omega_t$, and the algorithm is to determine for each data point from which of $\Omega_1, \ldots, \Omega_k$ it was chosen. In general we do *not* assume that $N_1, \ldots, N_t$ are revealed to the algorithm; but we do require some bounds on their relative sizes. An important parameter of the probability ensemble $\Omega_1, \ldots, \Omega_k$ is the *measure of divergence*

$$\gamma = \min_{1 \leq s < t \leq k} \frac{\sum_{i=1}^K (p_s^i - p_t^i)^2}{K} \tag{1.1}$$

between any two distributions. Note that $\sqrt{K\gamma}$ measures the Euclidean distance between the means of any two distributions and thus represents their separation. Further, let $N = n/k$ (so if the populations were balanced we would have $N$ of each type) and assume from now on that $kN < K$. Let $D = nK$ denote the size of the data-set. In addition, let $\sigma^2 = \max_{i,t} p_t^i(1 - p_t^i)$ denote the maximum variance of any random bit.

The biological context for this problem is we are given DNA information from $n$ individuals from $k$ populations of origin and we wish to classify each individual into the correct category. DNA contains a series of markers called SNPs, each of which has two variants (alleles). Given the population of origin of an individual, the genotypes can be reasonably assumed to be generated by drawing alleles independently from the appropriate distribution. The following theorem gives a sufficient condition for a balanced ($N_1 = N_2$) input instance when $k = 2$.

**Theorem 1.1** (Zhou (2006)). *Assume $N_1 = N_2 = N$. If $K = \Omega\left(\frac{\ln N}{\gamma}\right)$ and $KN = \Omega\left(\frac{\ln N \log \log N}{\gamma^2}\right)$ then with probability $1 - 1/\operatorname{poly}(N)$, among all balanced cuts in the complete graph formed among $2N$ sample individuals, the maximum weight cut corresponds to the partition of the $2N$ individuals according to their population of origin. Here the weight of a cut is the sum of weights across all edges in the cut, and the edge weight equals the Hamming distance between the bit vectors of the two endpoints.*

Variants of the above theorem, based on a model that allows two random draws from each SNP for an individual, are given in Chaudhuri et al. (2007); Zhou (2006). In particular, notice that edge weights based on the inner-product of two individuals' bit vectors correspond to the sample covariance, in which case the max-cut corresponds to the correct partition Zhou (2006) with high probability. Finding a max-cut is computationally intractable; hence in the same paper Chaudhuri et al. (2007), a hill-climbing algorithm is given to find the correct partition for balanced input instances but with a stronger requirement on the sizes of both $K$ and $nK$.

**A Spectral Approach:** In this paper, we construct two simpler algorithms using spectral techniques, attempting to reproduce conditions above. In particular, we study the requirements on the parameters of the model (namely, $\gamma$, $N$, $k$, and $K$) that allow us to classify every individual correctly and efficiently with high probability.

The two algorithms CLASSIFY and PARTITION compare as follows. Both algorithms are based on spectral methods originally developed in graph partitioning. More precisely, Theorem 1.2 is based on computing the singular vectors with the two largest singular values for each of the $n \times K$ input random matrix. The procedure is conceptually simple, easy to implement, and efficient in practice. For simplicity, Procedure Classify assumes the separation parameter $\gamma$ is known to decide which singular vector to examine; in practice, one can just try both singular vectors as we do in the simulations. Proof techniques for Theorem 1.2, however, are difficult to apply to cases of multiple populations, i.e., $k > 2$. Procedure Partition is based on computing a rank-$k$ approximation of the input random matrix and can cope with a mixture of a constant number of populations. It is more intricate for both implementation and execution than Classify. It does not require $\gamma$ as an input, while only requires that the constant $k$ is given. We prove the following theorems.

**Theorem 1.2.** *Let $\omega = \frac{\min(N_1, N_2)}{n}$ and $\omega_{\min}$ be a lower bound on $\omega$. Let $\gamma$ be given. Assume that $K > 2n \ln n$ and $k = 2$. Procedure* CLASSIFY *allows us to separate two populations w.h.p., when $n \geq \Omega\big(\frac{\sigma^2}{\gamma \omega_{\min} \omega}\big)$, where $\sigma^2$ is the largest variance of any random bit, i.e. $\sigma^2 = \max_{i,t} p_t^i(1 - p_t^i)$. Thus if the populations are roughly balanced, then $n \geq \frac{c}{\gamma}$ suffices for some constant $c$.*

This implies that the data required is $D = nK = O\left(\ln n \sigma^4 / \gamma^2 \omega^2 \omega_{\min}^2\right)$. Let $P_s = (p_s^i)_{i=1,\ldots,K}$, we have

$$\|P_1 - P_2\|_2 = \sqrt{K\gamma} = \sqrt{\sum_{i=1}^{K}(p_1^i - p_2^i)^2} \geq \frac{\sigma}{\omega_{\min}\omega}\sqrt{\ln n}. \tag{1.2}$$

**Theorem 1.3.** *Let $\omega = \frac{\min(N_1,\ldots,N_k)}{n}$. There is a polynomial time algorithm* PARTITION *that satisfies the following. Suppose that $K > n \log n$, $\gamma > K^{-2}$, $n > \frac{C_k \sigma^2}{\gamma \omega}$ for some large enough constant $C_k$, and $\omega = \Omega(1)$. Then given the empirical $n \times K$ matrix comprising the $K$ features for each of the $n$ individuals along with the parameter $k$,* PARTITION *separates the $k$ populations correctly w.h.p.*

**Summary and Future Direction:** Note that unlike Theorem 1.1, both Theorem 1.2 and Theorem 1.3 require a lower bound on $n$, even when $k = 2$ and the input instance is balanced. We illustrate through simulations to show that this seems not to be a fundamental constraint of the spectral techniques; our experimental results show that even when $n$ is small, by increasing $K$ so that $nK = \Omega(1/\gamma^2)$, one can classify a mixture of two populations using ideas in Procedure Classify with success rate reaching an "oracle" curve, which is computed assuming that distributions are known, where success rate means the ratio between correctly classified individuals and $N$. Exploring the tradeoffs of $n$ and $K$ that are sufficient for classification, when sample size $n$ is small, is both of theoretical interests and practical value.

**Outline of the paper:** The paper is organized as follows. In Section 1.1 we discuss related work. Then, in Section 2 we describe the algorithm CLASSIFY for Theorem 1.2 and outline its analysis. Some (very) technical details of the analysis are deferred to the appendix. Section 3 deals with the algorithm PARTITION for Theorem 1.3. Finally, in Section 4 we report some experimental results on CLASSIFY.

## *1.1. Related work*

In their seminal paper Pritchard et al. (2000), Pritchard, Stephens, and Donnelly presented a model-based clustering method to separate populations using genotype data. They assume that observations from each cluster are random from some parametric model. Inference for the parameters corresponding to each population is done jointly with inference for the cluster membership of each individual, and $k$ in the mixture, using Bayesian methods.

The idea of exploiting the eigenvectors with the first two eigenvalues of the adjacency matrix to partition graphs goes back to the work of Fiedler Fiedler (1973), and has been used in the heuristics for various NP-hard graph partitioning problems (e.g., Fjallstrom (1998)). The main difference between graph partitioning problems and the classification problem that we study is that the matrices occurring in graph partitioning are symmetric and hence diagonalizable, while our input matrix is rectangular in general. Thus, the contribution of Theorem 1.2 is to show that a conceptually simple and efficient algorithm based on singular value decompositions performs well in the framework of a fairly general probabilistic model, where probabilities for each of the $K$ features for each of the $k$ populations are allowed to vary. Indeed, the analysis of CLASSIFY requires exploring new ideas such as the Separation Lemma and the normalization of the random matrix $X$, for generating a large gap between top two singular values of the expectation matrix $\mathcal{X}$ and for bounding the angle between random singular vectors and their static correspondents, details of which are included in Section 2 with analysis in full version.

Procedure Partition and its analysis build upon the spectral techniques of McSherry (2001) on graph partitioning, and an extension due to Coja-Oghlan (2006). McSherry provides a comprehensive probabilistic model and presents a spectral algorithm for solving the partitioning problem on random graphs, provided that a separation condition similar to (1.2) is satisfied. Indeed, McSherry (2001) encompasses a considerable portion of the prior work on Graph Coloring, Minimum Bisection, and finding Maximum Clique. Moreover, McSherry's approach easily yields an algorithm that solves the classification problem studied in the present paper under similar assumptions as in Theorem 1.3, provided that the algorithm is given the parameter $\gamma$ as an additional input; this is actually pointed out in the conclusions of McSherry (2001). In the context of graph partitioning, an algorithm that does not need the separation parameter as an input was devised in Coja-Oghlan (2006). The main difference between PARTITION and the algorithm presented in Coja-Oghlan (2006) is that PARTITION deals with the asymmetric $n \times K$ matrix of individuals/features, whereas Coja-Oghlan (2006) deals with graph partitioning (i.e., a symmetric matrix).

There are two streams of related work in the learning community. The first stream is the recent progress in learning from the point of view of clustering: given samples drawn from a mixture of well-separated Gaussians (component distributions), one aims to classify each sample according to which component distribution it comes from, as studied in Dasgupta (1999), Dasgupta and Schulman (2000), Arora and Kannan (2001); Vempala and Wang (2002); Achlioptas and McSherry (2005); Kannan et al. (2005); Dasgupta et al. (2005). This framework has been extended to more general distributions such as log-concave distributions in Achlioptas and McSherry (2005); Kannan et al. (2005) and heavy-tailed distributions in Dasgupta et al. (2005), as well as to more than two populations. These results focus mainly on reducing the requirement on the separations between any two centers $P_1$ and $P_2$. In contrast, we focus on the sample size $D$. This is motivated by previous results Chaudhuri et al. (2007); Zhou (2006) stating that by acquiring enough attributes along the same set of dimensions from each component distribution, with high probability, we can correctly classify every individual.

While our aim is different from those results, where $n > K$ is almost universal and we focus on cases $K > n$, we do have one common axis for comparison, the $\ell_2$-distance between any two centers of the distributions. In earlier works Dasgupta and Schulman (2000); Arora and Kannan (2001), the separation requirement depended on the number of dimensions of each distribution; this has recently been reduced to be independent of $K$, the dimensionality of the distribution for certain classes of distributions Achlioptas and McSherry (2005); Kannan et al. (2005). This is comparable to our requirement in (1.2) for the discrete distributions. For example, according to Theorem 7 in Achlioptas and McSherry (2005), in order to separate the mixture of two Gaussians,

$$\|P_1 - P_2\|_2 = \Omega\left(\frac{\sigma}{\sqrt{\omega}} + \sigma\sqrt{\log n}\right) \tag{1.3}$$

is required. Besides Gaussian and Logconcave, a general theorem: Theorem 6 in Achlioptas and McSherry (2005) is derived that in principle also applies to mixtures of discrete distributions. The key difficulty of applying their theorem directly to our scenario is that it relies on a concentration property of the distribution (Eq. (10) of Achlioptas and McSherry (2005)) that need not hold in our case. In addition, once the distance between any two centers is fixed (i.e., once $\gamma$ is fixed in the discrete distribution), the sample size $n$ in their algorithms is always larger than $\Omega\left(\frac{K}{\omega}\log^5 K\right)$ Achlioptas and McSherry (2005); Kannan et al. (2005) for log-concave distributions (in fact, in Theorem 3 of Kannan et al. (2005), they discard at least this many individuals in order to correctly classify the rest in the sample), and larger than $\Omega(\frac{K}{\omega})$ for Gaussians Achlioptas and McSherry (2005), whereas in our case, $n < K$ always holds. Hence, our analysis allows one to obtain a clean bound on $n$ in the discrete case.

The second stream of work is under the PAC-learning framework, where given a sample generated from some target distribution $Z$, the goal is to output a distribution $Z_1$ that is close to $Z$ in Kullback-Leibler divergence: $KL(Z||Z_1)$, where $Z$ is a mixture of product distributions over discrete domains or Gaussians Kearns et al. (1994); Freund and Mansour (1999); Cryan (1999); Cryan et al. (2002); Mossel and Roch (2005); Feldman et al. (2005, 2006). They do not require a minimal distance between any two distributions, but they do not aim to classify every sample point correctly either, and in general require much more data.

Our work is also related to the use of *principal component analysis* ("PCA") in genetics Patterson et al. (2006); Price et al. (2006). The basic approach in these papers is to use the eigenvectors of a covariance matrix between samples to analyze a mixture of populations. While Patterson et al. (2006); Price et al. (2006) study the use of spectral methods empirically, the crucial point of the present work is that we prove rigorously that spectral methods succeed on a certain (simple) probabilistic model. Hence, our work can be seen as a further theoretical justification of the practical use of PCA. A difference between the present paper and Patterson et al. (2006); Price et al. (2006) is that we actually aim to assign each individual to exactly one of the populations. By contrast, Patterson et al. (2006); Price et al. (2006) just assign each individual a real "weight" for each population: essentially the eigenvectors with the dominant eigenvalues corresponding to the populations, and each individual is assigned its projection on these dominant eigenvectors. The algorithm CLASSIFY is somewhat similar to PCA, but PARTITION is conceptually more involved. In addition, our experimental results show a phase transition phenomenon similar to what was observed in Patterson et al. (2006) in detecting population structure using simulated data.

## 2. A simple algorithm using singular vectors

As described in Theorem 1.2, we assume we have a mixture of two product distributions. Let $N_1, N_2$ be the number of individuals from each population class. Our goal is to correctly classify all individuals according to their distributions. Let $n = 2N = N_1 + N_2$, and refer to the case when $N_1 = N_2$ as the balanced input case. For convenience, let us redefine "$K$" to assume we have $O(\log n)$ blocks of $K$ features each (so the total number of features is really $O(K \log n)$) and we assume that each set of $K$ features has divergence at least $\gamma$. (If we perform this partitioning of features into blocks randomly, then with high probability this divergence has changed by only a constant factor for most blocks.)

The high-level idea of the algorithm is now to repeat the following procedure for each block of $K$ features: use the $K$ features to create an $n \times K$ matrix $X$, such that each row $X_i, i = 1, \ldots, n$, corresponds to a feature vector for one sample point, across its $K$ dimensions. We then compute the top two left singular vectors $u_1, u_2$ of $X$ and use these to classify each sample. This classification induces some probability of error $f$ for each individual at each round, so we repeat the procedure for each of the $O(\log n)$ blocks and then take majority vote over different runs. Each round we require $K \geq n$ features, so we need $O(n \log n)$ features total in the end.

In more detail, we repeat the following procedure $O(\log n)$ times. Let $T = \frac{15N}{32}\sqrt{3\omega_{\min}\gamma}$, where $\omega_{\min}$ is the lower bound on the minimum weight $\min\{\frac{N_1}{2N}, \frac{N_2}{2N}\}$, which is independent of an actual instance. Let $s_1(X), s_2(X)$ be the top two singular values of $X$.

**Procedure Classify:** Given $\gamma, N, \omega_{\min}$. Assume that $N \gg \frac{1}{\gamma}$,

- Normalization: use the $K$ features to form a random $n \times K$ matrix $X$; Each individual random variable $X_{i,j}$ is a *normalized* random variable based on the original Bernoulli r.v. $b_{i,j} \in \{0, 1\}$ with $\mathbf{Pr}[b_{i,j} = 1] = p_1^j$ for $X_i \in P_1$ and $\mathbf{Pr}[b_{i,j} = 1] = p_2^j$ for $X_i \in P_2$, such that $X_{i,j} = \frac{b+1}{2}$.
- Take top two left singular vectors $u_1, u_2$ of $X$, where $u_i = [u_{i,1}, \ldots, u_{i,n}], i = 1, 2$.
  1. If $s_2(X) > T = \frac{15N}{32}\sqrt{3\omega_{\min}\gamma}$, use $u_2$ to partition the individuals with 0 as the threshold, i.e., partition $j \in [n]$ according to $u_{2,j} < 0$ or $u_{2,j} \geq 0$.
  2. Otherwise, use $u_1$ to partition, with mixture mean $M = \sum_{i=1}^{n} u_{1,n}$ as the threshold.

**Analysis of the Simple Algorithm:** Our analysis is based on comparing entries in the top two singular vectors of the normalized random $n \times K$ matrix $X$, with those of a static matrix $\mathcal{X}$, where each entry $\mathcal{X}_{i,j} = \mathbf{E}[X_{i,j}]$ is the expected value of the corresponding entry in $X$. Hence $\forall i = 1, \ldots, N_1$, $\mathcal{X}_i = [\mu_1^1, \mu_1^2, \ldots, \mu_1^K]$, where $\mu_1^j = \frac{1+p_1^j}{2}, \forall j$, and $\forall i = N_1 + 1, \ldots, n$, $\mathcal{X}_i = [\mu_2^1, \mu_2^2, \ldots, \mu_2^K]$, where $\mu_2^j = \frac{1+p_2^j}{2}, \forall j$. We assume the divergence is exactly $\gamma$ among the $K$ features that we have chosen in all calculations.

The inspiration for this approach is based on Lemma 2.1, whose proof is built upon Theorem A.2 that is presented in a lecture note by Spielman (2002). For an $n \times K$ matrix $A$, let $s_1(A) \geq s_2(A) \geq \cdots \geq s_n(A)$ be singular values of $A$. Let $u_1, \ldots, u_n, v_1, \ldots, v_n$, be the $n$ left and right singular vectors of $X$, corresponding to $s_1(X), \ldots, s_n(X)$ such that $\|u_i\|_2 = 1, \|v_i\|_2 = 1, \forall i$. We denote the set of $n$ left and right singular vectors of $\mathcal{X}$ with $\bar{u}_1, \ldots, \bar{u}_n, \bar{v}_1, \ldots, \bar{v}_n$.

**Lemma 2.1.** *Let $X$ be the random $n \times K$ matrix and $\mathcal{X}$ its expected value matrix. Let $A = X - \mathcal{X}$ be the zero-mean random matrix. Let $\theta_i$ be the angle between two vectors: $[u_i, v_i], [\bar{u}_i, \bar{v}_i]$, where $\|[u_i, v_i]\|_2 = \|[\bar{u}_i, \bar{v}_i]\|_2 = 2$ and $[u, v]$* *represents a vector that is the concatenation of two vectors $u, v$.*

$$\|u_i - \bar{u}_i\|_2 \leq \|[u_i, v_i] - [\bar{u}_i, \bar{v}_i]\|_2 \approx 2\theta_i \approx 2\sin(\theta_i) \leq \frac{4s_1(A)}{\mathsf{gap}(i, \mathcal{X})}, \qquad (2.1)$$

where $\mathsf{gap}(i, \mathcal{X}) = \min_{j \neq i} |s_i(\mathcal{X}) - s_j(\mathcal{X})|$.

We first bound the largest singular value $s_1(A) = s_1(X - \mathcal{X})$ of $(a_{i,j})$ with independent zero-mean entries, which defines the Euclidean operator norm

$$\|(a_{i,j})\| := \sup\left\{\sum_{i,j} a_{i,j} x_i y_j : \sum x_i^2 \leq 1, \sum y_i^2 \leq 1\right\}. \qquad (2.2)$$

The behavior of the largest singular value of an $n \times m$ random matrices $A$ with i.i.d. entries is well studied. Latala (2005) shows that the weakest assumption for its regular behavior is boundedness of the fourth moment of the entries, even if they are not identically distributed. Combining Theorem 2.2 of Latala (2005) with the concentration Theorem 2.3 by Meckes (2004) proves Theorem 2.4 that we need [1].

**Theorem 2.2** (Norm of Random Matrices Latala (2005)). *For any finite $n \times m$ matrix $A$ of independent mean zero r.v.'s $a_{i,j}$ we have, for an absolute constant $C$,*

$$\mathbf{E}\|(a_{i,j})\| \leq C\left(\max_i \sqrt{\sum_j \mathbf{E} a_{i,j}^2} + \max_j \sqrt{\sum_i \mathbf{E} a_{i,j}^2} + \left(\sum_{i,j} \mathbf{E} a_{i,j}^4\right)^{\frac{1}{4}}\right). \qquad (2.3)$$

**Theorem 2.3** (Concentration of Largest Singular Value: Bounded Range Meckes (2004)). *For any finite $n \times m$, where $n \leq m$, matrix $A$, such that entries $a_{i,j}$ are independent r.v. supported in an interval of length at most $D$, then, for all $t$,*

$$\mathbf{Pr}[|s_1(A) - \mathbb{M}s_1(A)| \geq t] \leq 4e^{-t^2/4D^2}. \qquad (2.4)$$

**Theorem 2.4** (Largest Singular Value of a Mean-zero Random Matrix). *For any finite $n \times K$, where $n \leq K$, matrix $A$, such that entries $a_{i,j}$ are independent mean zero r.v. supported in an interval of length at most $D$, with fourth moment upper bounded by $B$, then*

$$\mathbf{Pr}\left[s_1(A) \geq CB^{1/4}\sqrt{K} + 4D\sqrt{\pi} + t\right] \leq 4e^{-t^2/4} \qquad (2.5)$$

*for all $t$. Hence $\|A\| \leq C_1 B^{1/4}\sqrt{K}$ for an absolute constant $C_1$.*

---

[1] One can also obtain an upper bound of $O(\sqrt{n+K})$ on $s_1(A)$ using a theorem on by Vu (2005), through the construction a $(n+K) \times (n+K)$ square matrix out of $A$.

### 2.1. Generating a large gap in $s_1(\mathcal{X}), s_2(\mathcal{X})$

In order to apply Lemma 2.1 to the top two singular vectors of $X$ and $\mathcal{X}$ through

$$\|u_1 - \bar{u}_1\|_2 \leq \frac{4s_1(X - \mathcal{X})}{|s_1(\mathcal{X}) - s_2(\mathcal{X})|} \tag{2.6}$$

$$\|u_2 - \bar{u}_2\|_2 \leq \frac{4s_1(X - \mathcal{X})}{\min\left(|s_1(\mathcal{X}) - s_2(\mathcal{X})|, |s_2(\mathcal{X})|\right)}, \tag{2.7}$$

we need to first bound $|s_1(\mathcal{X}) - s_2(\mathcal{X})|$ away from zero, since otherwise, RHSs on both (2.6) and (2.7) become unbounded. We then analyze

$$\mathsf{gap}(2, \mathcal{X}) = \min\left(|s_1(\mathcal{X}) - s_2(\mathcal{X})|, |s_2(\mathcal{X})|\right).$$

Let us first define values $a, b, c$ that we use throughout the rest of the paper:

$$a = \sum_{k=1}^{K} (\mu_1^k)^2, \quad b = \sum_{k=1}^{K} \mu_1^k \mu_2^k, \quad c = \sum_{k=1}^{K} (\mu_2^k)^2. \tag{2.8}$$

For the following analysis, we can assume that $a, b, c \in [K/4, K]$, given that $X$ is normalized in Procedure Classify.

We first show that normalization of $X$ as described in Procedure Classify guarantees that not only $|s_1(\mathcal{X}) - s_2(\mathcal{X})| \neq 0$, but there also exists a $\Theta(\sqrt{NK})$ amount of gap between $s_1(\mathcal{X})$ and $s_2(\mathcal{X})$ in Proposition 2.5:

$$\mathsf{gap}(\mathcal{X}) := |s_1(\mathcal{X}) - s_2(\mathcal{X})| = \Theta(\sqrt{NK}). \tag{2.9}$$

**Proposition 2.5.** *For a normalized random matrix $X$, its expected value matrix $\mathcal{X}$ satisfies $\frac{4c_0\sqrt{2NK}}{5} \leq \mathsf{gap}(\mathcal{X}) \leq \sqrt{2NK}$, where $c_0 = \frac{|b|\sqrt{ac}}{K(a+c)}$ is a constant, given that $a, b, c \in [K/4, K]$ as defined in (2.8). In addition,*

$$\sqrt{\frac{KN}{4}} \leq s_1(\mathcal{X}) \leq \sqrt{2NK}, \text{ and } \sqrt{\frac{NK}{2}} \leq s_1(\mathcal{X}) + s_2(\mathcal{X}) \leq \sqrt{2NK}. \tag{2.10}$$

We next state a few important results that justify Procedure Classify. Note that the left singular vectors $\bar{u}_i, \forall i$ of $\mathcal{X}$ are of the form $[x_i, \ldots, x_i, y_i, \ldots, y_i]^T$:

$$\bar{u}_1 = [x_1, \ldots, x_1, y_1, \ldots, y_1]^T, \text{ and } \bar{u}_2 = [x_2, \ldots, x_2, y_2, \ldots, y_2]^T, \tag{2.11}$$

where $x_i$ repeats $N_1$ times and $y_i$ repeats $N_2$ times. We first show Proposition 2.6 regarding signs of $x_i, y_i, i = 1, 2$, followed by a lemma bounding the separation of $x_2, y_2$. We then state the key Separation Lemma that allows us to conclude that least one of top two left singular vectors of $X$ can be used to classify data at each round. It can be extended to cases when $k > 2$.

**Proposition 2.6.** *Let $b$ as defined in (2.8): when $b > 0$, entries $x_1, y_1$ in $\bar{u}_1$ have the same sign while $x_2, y_2$ in $\bar{u}_2$ have opposite signs.*

**Lemma 2.7.** $|x_2 - y_2|^2 \leq \frac{C_{\max}}{2N}$ where $C_{\max} = \left(\sqrt{\frac{1}{\omega_1}} + \sqrt{\frac{1}{\omega_2}}\right)^2 \leq \frac{4}{\omega_{\min}}$; $|x_2|^2 \geq \frac{C_{x\min}}{2N}$ where $C_{x\min} = \frac{\omega_2}{4\omega_1^2+\omega_1\omega_2}$; $|y_2|^2 \geq \frac{C_{y\min}}{2N}$ where $C_{y\min} = \frac{\omega_1}{4\omega_2^2+\omega_1\omega_2}$.

**Lemma 2.8** (Separation Lemma). $K\gamma = s_1(\mathcal{X})^2(x_1-y_1)^2 + s_2(\mathcal{X})^2(x_2-y_2)^2$.

*Proof.* Let $\Delta := P_1 - P_2$ as in Theorem 1.2, and $\vec{b} = [1, 0, \ldots, 0, -1, 0, \ldots, 0]^T$, where 1 appears in the first and $-1$ appears in the $N_1 + 1^{st}$ positions. Then $\Delta = X^T\vec{b} = [\mu_1^1 - \mu_2^1, \mu_1^2 - \mu_2^2, \ldots, \mu_1^K - \mu_2^K]$. Given $\mathcal{X} = s_1(\mathcal{X})\bar{u}_1\bar{v}_1^T + s_2(\mathcal{X})\bar{u}_2\bar{v}_2^T$, we thus rewrite $\Delta$ as: $\Delta = \mathcal{X}^T\vec{b} = s_1(\mathcal{X})\bar{v}_1\bar{u}_1^T\vec{b} + s_2(\mathcal{X})\bar{v}_2\bar{u}_2^T\vec{b} = s_1(\mathcal{X})\bar{v}_1(x_1-y_1) + s_2(\mathcal{X})\bar{v}_2(x_2-y_2)$. The lemma follows from the fact that $\|\Delta\|_2 = \sqrt{K\gamma}$ and $\bar{v}_1, \bar{v}_2$ are orthonormal. □

Combining Proposition 2.6, Lemma 2.7, (2.10), and Lemma 2.8, we have

**Corollary 2.9.** $s_2(\mathcal{X}) \leq \frac{\sqrt{2NK\gamma}}{\sqrt{c_{x\min}}+\sqrt{c_{y\min}}}$, and hence $\mathsf{gap}(2,\mathcal{X}) = \min(s_2(\mathcal{X}), |s_1(\mathcal{X}) - s_2(\mathcal{X})|) = s_2(\mathcal{X})$ for a sufficiently small $\gamma$.

In Section 2.2, we first prove a proposition regarding $a, b, c$ as defined in (2.8). We next provide the proof for Theorem 2.4 regarding the largest singular value of $(X - \mathcal{X})$. In Section 2.4, we show that the probability of error at each round for each individual is at most $f = 1/10$, given the sample size $n$ as specified in Theorem 1.2. Hence by taking majority vote over the different runs for each sample, our algorithm will find the correct partition with probability $1 - 1/n^2$, given that at each round we take a set of $K > n$ independent features.

## 2.2. Detailed analysis for the simple algorithm

Throughout the rest of the paper, we use $X, Y, H$ to represent random matrices, where $H = XX^T$ and $Y = \begin{bmatrix} 0 & X \\ X^T & 0 \end{bmatrix}$. We use $\mathcal{X}, \mathcal{Y}, \mathcal{H}$ to represent the corresponding static matrices. Let us substitute $a, b, c$ in $\mathcal{H} = \mathcal{X}\mathcal{X}^T$, where the blocks in $\mathcal{H}$ from top to bottom and from left to right are of size: $N_1 \times N_1, N_1 \times N_2, N_2 \times N_1$ and $N_2 \times N_2$ respectively:

$$\mathcal{H} = \mathcal{X}\mathcal{X}^T = \begin{bmatrix} a & \ldots & a & b & \ldots & b \\ \ldots & & & & & \\ a & \ldots & a & b & \ldots & b \\ b & \ldots & b & c & \ldots & c \\ \ldots & & & & & \\ b & \ldots & b & c & \ldots & c \end{bmatrix}_{2N \times 2N}. \quad (2.12)$$

**Proposition 2.10.** *For any choices of $\mu_i^k$, $ac \geq b^2$; By definition,*

$$a + c - 2b = \sum_{i=1}^{K} \alpha_k^2, \text{ where } \alpha_k = |\mu_1^k - \mu_2^k|. \quad (2.13)$$

*Proof.* $a + c - 2b = \sum_k \alpha_k^2$ holds by definition.

$$\begin{aligned}
ac - b^2 &= \sum_{k=1}^K (\mu_1^k)^2 \sum_{k=1}^K (\mu_2^k)^2 - \left(\sum_{k=1}^K (\mu_1^k \mu_2^k)\right)^2 \\
&= \sum_{k=1}^K (\mu_1^k \mu_1^k)^2 + \sum_{j \neq k} ((\mu_1^k \mu_2^j)^2 + (\mu_1^j \mu_2^k)^2) \\
&\quad - \left(\sum_{k=1}^K (\mu_1^k \mu_1^k)^2 + \sum_{j \neq k} 2\mu_1^k \mu_2^k \mu_1^j \mu_2^j \right) \\
&= \sum_{j \neq k} (\mu_1^k \mu_2^j)^2 + (\mu_1^j \mu_2^k)^2 - 2\mu_1^k \mu_2^k \mu_1^j \mu_2^j = \sum_{j \neq k} (\mu_1^k \mu_2^j - \mu_1^j \mu_2^k)^2 \geq 0.
\end{aligned}$$

□

**Remark 2.11.** *Both matrices of $\mathcal{X}$ and $\mathcal{X}\mathcal{X}^T$ have rank at most two. When $ac = b^2$, $\mathcal{H}$ has rank 1.*

### 2.3. Proof of Theorem 2.4

By having an upper bound on both maximum variance and fourth moment of any entry, we have the following corollary of Theorem 2.2.

**Corollary 2.12** (Largest Singular Value: Bounded Fourth Moment Latala (2005)). *For any finite $n \times m$, where $n \leq m$, matrix of independent mean zero r.v.'s $a_{i,j}$, such that the maximum variances of any entry is at most $\sigma^2$, and each entry has a finite fourth moment $B$ we have*

$$\mathbf{E}\|(a_{i,j})\| \leq C\left(\sigma(\sqrt{m} + \sqrt{n}) + (mnB)^{1/4}\right) \leq CB^{1/4}\sqrt{m} \tag{2.14}$$

*for an absolute constant $C$.*

**Remark 2.13.** *The requirement that $\sigma^2$ is upper bounded is not essential. The conclusion in Corollary 2.12 works so long as fourth moment is bounded by $B$.*

Let $\mathbb{M}s_1(A)$ be the median of $s_1(A)$. Following a calculation from Meckes (2004), we have

$$\begin{aligned}
|\mathbf{E}[s_1(A)] - \mathbb{M}s_1(A)| &\leq \mathbf{E}[|s_1(A) - \mathbb{M}s_1(A)|] \\
&= \int_0^\infty \mathbf{Pr}[|s_1(A) - \mathbb{M}s_1(A)| \geq t] dt \\
&\leq 4\int_0^\infty e^{-t^2/4D^2} dt = 4D\sqrt{\pi},
\end{aligned}$$

where $D \leq 1$ for Bernoulli random variables that we consider. This allows us to conclude Theorem 2.4. □

### *2.4. Correctness of classification for the simple algorithm*

We now prove correctness of our algorithm. We first show how to choose $T$ for Procedure Classify. Let $B$ denote the fourth moment bound for a single random variable in the mean zero random matrix $X - \mathcal{X}$; for the type of normalized Bernoulli r.v.s that we care about, $\sqrt{B}$ is in the order of $\sigma^2$, where $\sigma^2$ is defined in Theorem 1.2.

Let $N\gamma$ be a large enough constant. Let $s_1(X - \mathcal{X}) \leq C_0\sqrt{K}$, where $C_0 = C_1 B^{1/4}$ as defined in Theorem 2.4 and let the threshold

$$T = \sqrt{C_3 K N\gamma} \geq 15 C_0 \sqrt{K}, \tag{2.15}$$

which requires that

$$C_3 N\gamma \geq 225 C_0^2, \text{ where } C_3 \text{ satisfies } (2.19). \tag{2.16}$$

Following Lemma A.4, (A.1), (A.3), and Proposition A.1, we have

$$|s_2(\mathcal{X}) - s_2(X)| \leq s_1(X - \mathcal{X}) \leq C_0\sqrt{K}. \tag{2.17}$$

We have two cases,

1. When $s_2(X) \leq T$, by Lemma 2.7 and the fact that $s_2(\mathcal{X}) \leq s_2(X) + s_1(X - \mathcal{X}) \leq T + C_0\sqrt{K} \leq \frac{16T}{15}$, we have

$$s_2(\mathcal{X})^2 |x_2 - y_2|^2 \leq \frac{256 T^2}{225} \frac{C_{\max}}{2N} \leq \frac{128 C_3 K\gamma C_{\max}}{225} \leq \frac{128 C_3 K \gamma 4}{225 \omega_{\min}} \tag{2.18}$$

for $C_{\max}$ as defined in Lemma 2.7. We want $s_2(\mathcal{X})^2 |x_2 - y_2|^2 \leq \frac{3K\gamma}{4}$. This holds so long as $\frac{128 C_3 K\gamma C_{\max}}{225} \leq \frac{128 C_3 K\gamma 4}{225 \omega_{\min}} \leq \frac{3K\gamma}{4}$, which is true if

$$C_3 \leq \frac{675 \omega_{\min}}{2048}; \text{ thus we take } C_3 = \frac{675 \omega_{\min}}{2048} \text{ from this point on.} \tag{2.19}$$

It follows from Lemma 2.8 that $s_1(\mathcal{X})^2 |x_1 - y_1|^2 \geq \frac{K\gamma}{4}$. Hence by (2.10)

$$|x_1 - y_1| \geq \frac{\sqrt{K\gamma}}{2 s_1(\mathcal{X})} \geq \frac{\sqrt{K\gamma}}{2\sqrt{2NK}} \geq \frac{1}{2}\sqrt{\frac{\gamma}{2N}}. \tag{2.20}$$

Thus the condition of Theorem 2.14 holds with $c_2 = \frac{1}{2}$, so long as

$$N\gamma \geq \frac{2048 C_1^2 \sqrt{B}}{3\omega_{\min}}, \tag{2.21}$$

due to (2.16) and (2.19); This is a weaker condition than (2.32) for $f < \frac{1}{2}$.
2. When $s_2(X) \geq T$, we have $s_2(\mathcal{X}) \geq s_2(X) - s_1(X - \mathcal{X}) \geq T - C_0\sqrt{K} \geq \frac{14T}{15}$; This satisfies the condition of Theorem 2.18, with $c_3 = \frac{14\sqrt{C_3}}{15} = \frac{7}{16}\sqrt{\frac{3\omega_{\min}}{2}}$.

Let us first denote the first singular vector $u_1$ and its "noise" vector $\epsilon$ as follows:

$$u_1^T = (x + \delta_1, \ldots, x + \delta_{N_1}, y + \tau_1, \ldots, y + \tau_{N_2}), \ \epsilon^T = (\delta_1, \ldots, \delta_{N_1}, \tau_1, \ldots, \tau_{N_2}).$$

It turns out that we only need to use the mixture mean

$$M = \frac{\sum_{i=1}^{N_1}(x + \delta_i) + \sum_{i=1}^{N_2}(y + \tau_i)}{2N} \tag{2.22}$$

to decide which side to put a node, i.e., to partition $j \in [2N]$ according to $u_{1,j} < M$ or $u_{1,j} \geq M$, given that $N_1/N_2$ is a constant; Misclassifying any entry will contribute $\Omega\left(\frac{\gamma}{2N}\right)$ amount to $\|\bar{u}_1 - u_1\|_2^2$.

**Theorem 2.14.** *Assume w.l.o.g. that $N_1 \leq N_2$ and $2N \leq K$. Let $\omega_1 = N_1/2N$ and $\omega_2 = N_2/2N$. Suppose $|x_1 - y_1| \geq c_2\sqrt{\frac{\gamma}{2N}}$ for some constant $c_2 = \frac{1}{2}$. By requiring $N \geq \frac{2048 C_1^2 \sqrt{B}}{3\gamma \omega_{\min}}$ as in (2.21), and*

$$N_1 \geq \frac{2c_1^2 \sigma^2}{f c_2^2 \gamma \omega_1 \omega_2}, \quad \text{or equivalently} \quad 2N \geq \frac{2c_1^2 \sigma^2}{f c_2^2 \gamma \omega_2 \omega_1^2} = \frac{25 C_1^2 \sqrt{B}}{f c_2^2 \gamma \omega_2 \omega_1^2}, \tag{2.23}$$

*where $c_1 = \frac{5 C_1 B^{1/4}}{\sqrt{2} c_0 \sigma}$ for $C_1$ specified in Theorem 2.4 and $c_0$ specified in Proposition 2.5, we can classify the two population using the mixture mean $M$ with the error factor at most $f$ for $N_1, N_2$ respectively whp.*

By Lemma 2.1 and Theorem 2.4, we immediately have the following claim.

**Claim 2.15.** *For $c_1$ chosen as in Theorem 2.14, $\|\epsilon\|_2^2 = \sum_{i=1}^{N_1} \delta_i^2 + \sum_{i=1}^{N_2} \tau_i^2 \leq \frac{c_1^2 \sigma^2}{N}$.*

*Proof.* Given that $c_1 = \frac{5 C_1 \sqrt[4]{B}}{\sqrt{2} c_0 \sigma}$ such that $C_1$ appears in Theorem 2.4 and $c_0$ appears in Proposition 2.5,

$$\sqrt{\sum_{i=1}^{N_1} \delta_i^2 + \sum_{i=1}^{N_2} \tau_i^2} = \|u_1 - \bar{u}_1\|_2 \sim 2\theta_1 \sim 2\sin(\theta_1)$$

$$\leq \frac{4 s_1(X - \mathcal{X})}{\mathsf{gap}(1, \mathcal{X})} \leq \frac{4 C_1 \sqrt[4]{B} \sqrt{K}}{4 c_0 \sqrt{2NK}/5} = \frac{c_1 \sigma}{\sqrt{N}}.$$

This allows us to conclude the claim. $\square$

We need the following lemma, proof of which appears in Appendix B.

**Lemma 2.16.** *Assume that $2N \leq K$ and Condition (2.23) in Theorem 2.14, we have*

$$|M - x| \geq \frac{N_2(1 - \sqrt{\gamma})|y - x|}{2N}, \quad |y - M| \geq \frac{N_1(1 - \sqrt{\gamma})|y - x|}{2N}. \tag{2.24}$$

*Proof of Theorem 2.14.* Recall that the largest $\bar{u}_1, \bar{u}_2$ have the form of $[x, \ldots, x, y, \ldots, y]$, where $x$ repeats $N_1$ times and $y$ repeats $N_2$ times; hence w.l.o.g., assume that $x < y$, we have

$$\forall i, s.t. \ x + \delta_i > M, \text{ it contributes } \delta_i^2 > |M - x|^2 \geq \approx \frac{N_2^2 c_2^2 \gamma}{8N^3} \text{ to } \|\epsilon\|_2^2, \quad (2.25)$$

$$\forall i, s.t. \ y - \tau_i < M, \text{ it contributes } \delta_i^2 > |M - y|^2 \geq \approx \frac{N_2^2 c_2^2 \gamma}{8N^3} \text{ to } \|\epsilon\|_2^2. \quad (2.26)$$

Hence the total number of entries that goes above $M$ from $P_1$, and those goes below $M$ from $P_2$ can not be too many since their total contribution is upper bounded by $\|\epsilon\|_2^2 = \|u_1 - \bar{u}_1\|_2^2$. Let $\ell_1$ be the number of misclassified entries from $N_1$, i.e., those described in (2.25), by Lemma 2.1,

$$\ell_1 \frac{N_2^2 c_2^2 \gamma}{8N^3} \leq \ell_1 |M - x|^2 \leq \|\epsilon\|_2^2 \leq \frac{c_1^2 \sigma^2}{N}. \quad (2.27)$$

Thus given that $N_1 \geq \frac{8c_1^2 \sigma^2}{fc_2^2 \gamma} \geq \frac{2c_1^2 \sigma^2}{f\omega_2^2 c_2^2 \gamma}$; hence it suffices to guarantee that $\ell_1 \leq \frac{2c_1^2 \sigma^2}{\omega_2^2 c_2^2 \gamma} \leq fN_1$.

We next bound the number of entries from $P_2$ that goes below $M$, which can not be too many either; let $\ell_2$ be the number of misclassified entries from $P_2$,

$$\ell_2 \frac{N_1^2 c_2^2 \gamma}{8N^3} \leq \ell_2 |M - y|^2 \leq \|\epsilon\|_2^2 \leq \frac{c_1^2 \sigma^2}{N}, \quad (2.28)$$

hence by requiring

$$N_2 \geq \frac{2c_1^2 \sigma^2}{f\omega_1^2 c_2^2 \gamma}, \quad (2.29)$$

it suffices to guarantee that $\ell_2 \leq \frac{2c_1^2 \sigma^2}{\omega_2^2 c_2^2 \gamma} \leq fN_2$.

Condition (2.29) is equivalent to

$$N_1 = \frac{N_2 \omega_1}{\omega_2} \geq \frac{2c_1^2 \sigma^2}{f\omega_1 \omega_2 c_2^2 \gamma}, \quad (2.30)$$

Thus by requiring

$$N_1 \geq \frac{2c_1^2 \sigma^2}{fc_2^2 \gamma \omega_2 \omega_1}, \quad (2.31)$$

we have satisfied all requirements. □

Combining Lemma 2.1 and Corollary 2.9, we have

**Claim 2.17.** *Given that $s_2(\mathcal{X}) \geq c_3 \sqrt{KN\gamma}$, $\|u_2 - \bar{u}_2\|_2^2 \leq \frac{16 s_1 (X - \mathcal{X})^2}{s_2(\mathcal{X})^2} \leq \frac{16 C_0^2 K}{c_3^2 KN\gamma}$.*

This allows us to prove the following theorem. Let the classification error factor be the number of misclassified individuals from one group over total amount of people in that group.

**Theorem 2.18.** *Assume $N_1 \leq N_2$ and $2N \leq K$. Let $\omega_1 = N_1/2N$ and $\omega_2 = N_2/2N$. Let $s_2(\mathcal{X}) \geq c_3\sqrt{KN\gamma}$, where $c_3 = \frac{7}{16}\sqrt{\frac{3\omega_{\min}}{2}}$ and $\omega_{\min}$ is the minimum possible weight allowed by the algorithm. By requiring*

$$2N \geq \frac{360 C_0^2}{\omega_{\min} f \gamma}\left(\frac{\omega_2}{\omega_1} + 1\right) = \Theta\left(\frac{\sigma^2}{f\gamma\omega_{\min}\omega_1}\right), \quad (2.32)$$

*we can classify the two population using $0$ to separate components in $u_2$, with error factor at most $f$ for both $P_1, P_2$ whp.*

*Proof.* Let $\ell_1, \ell_2$ be the number of misclassified entries from $P_1$ and $P_2$ respectively; they each contribute at least $\frac{C_{x\min}}{2N}$, and $\frac{C_{y\min}}{2N}$ amount to $\|u_2 - \bar{u}_2\|_2$, and hence by Claim 2.17,

$$\ell_1 \frac{C_{x\min}}{2N} \leq \|u_2 - \bar{u}_2\|_2^2 \leq \frac{16 C_0^2 K}{c_3^2 K N \gamma} \leq \frac{16 C_0^2}{c_3^2 N \gamma}. \quad (2.33)$$

Hence $\ell_1 \leq \frac{32 C_0^2}{c_3^2 \gamma C_{x\min}} \leq f N_1$ given that $N \geq \frac{16 C_0^2}{c_3^2 f \gamma}(4\frac{\omega_1}{\omega_2} + 1)$.

Similarly, by Claim 2.17, we have $\ell_2 \frac{C_{x\min}}{2N} \leq \|u_2 - \bar{u}_2\|_2^2$ and thus $\ell_2 \leq \frac{32 C_0^2}{c_3^2 \gamma C_{y\min}} \leq \frac{N_2}{f}$ so long as $N \geq \frac{16 C_0^2}{c_3^2 f \gamma}(4\frac{\omega_2}{\omega_1} + 1)$; the bound on $2N$ follows by plugging in $c_3 = \frac{7}{16}\sqrt{\frac{3\omega_{\min}}{2}}$. □

Finally,

**Theorem 2.19.** *Given a set of $n \geq \Omega\left(\frac{\sigma^2}{\gamma f \omega \omega_{\min}}\right)$ individuals, by trying Procedure Classify for $\log n$ rounds, with probability of error at each round for each individual being $f = 1/10$, where each round we take a set of $K > n$ independent features, and by taking majority vote over the different runs for each sample, our algorithm will find the correct partition with probability $1 - 1/n^2$.*

*Proof.* A sample is put in the wrong side with a probability $1/10$ at each round. Let $\mathcal{E}_i$ be the event that sample $i$ is misclassified for more than $\log n$ times, thus $\mathbf{Pr}[\mathcal{E}_i] = \left(\frac{1}{10}\right)^{\log n} \leq 1/n^{3.32}$; hence by union bound, with probability $1 - 1/n^2$, none of the $2N$ individuals is misclassified. □

## 3. The algorithm PARTITION

### 3.1. Preliminaries

Let $V = \{1, \ldots, n\}$ be the set of all $n$ individuals, and let $\psi : V \to \{1, \ldots, k\}$ be the map that assigns to each individual the population it belongs to. Set $V_t = \psi^{-1}(t)$ and $N_t = |V_t|$. Moreover, let $\mathbb{E} = (\mathbb{E}_{vi})_{1 \leq v \leq n, 1 \leq i \leq K}$ be the $n \times K$ matrix with entries $\mathbb{E}_{vi} = p^i_{\psi(v)}$. For any $1 \leq t \leq k$ we let $\mathbb{E}^{V_t} = (p^l_t)_{l=1,\ldots,K}$ be the row of $\mathbb{E}$ corresponding to any $v \in V_t$. In addition, let $A = (a_{vi})$ denote the empirical $n \times K$ input matrix. Thus, the entries of $\mathbb{E}$ equal the expectations of the entries of $A$.

As in Theorem 1.3, we let

$$\gamma = K^{-1} \min_{1 \leq i < j \leq k} \|\mathbb{E}^{V_i} - \mathbb{E}^{V_j}\|^2, \quad \Gamma = K\gamma.$$

Further, set $\lambda = \sqrt{K}\sigma$. Then the assumption from Theorem 1.3 can be rephrased as

$$n_{\min}\Gamma > C_k \lambda^2 \text{ and } \Gamma > K^{-1} \tag{3.1}$$

where $C_k$ signifies a sufficiently large number that depends on $k$ only (the precise value of $C_k$ will be specified implicitly in the course of the analysis). As in the previous section, by repeating the partitioning process $\log n$ times, we may restrict our attention to the problem of classifying a constant fraction of the individuals correctly. That is, it is sufficient to establish the following claim.

**Claim 3.1.** *There is a polynomial time algorithm* `Partition` *that satisfies the following. Suppose that (3.1) is true. Then whp* `Partition(A,k)` *outputs a partition* $(S_1, \ldots, S_k)$ *of $V$ such that there exists a permutation $\sigma$ such that*

$$\sum_{i=1}^{k} |V_i \triangle S_{\sigma(i)}| < 0.001 n_{\min}.$$

Let $X = (x_{ij})_{1 \leq i \leq n, 1 \leq j \leq K}$ be a $n \times K$ matrix. By $X_i$ we denote the $i$'th row $(X_{i1}, \ldots, X_{iK})$ of $X$. Moreover, we let

$$\|X\| = \max_{\xi \in \mathbf{R}^K : \|\xi\|=1} \|X\xi\|$$

signify the *operator norm* of $X$. A *rank $k$ approximation* of $X$ is a matrix $\widehat{X}$ of rank at most $k$ such that for any $n \times K$ matrix $Y$ of rank at most $k$ we have $\|X - \widehat{X}\| \leq \|X - Y\|$. Given $X$, a rank $k$ approximation $\widehat{X}$ can be computed as follows. Letting $\rho = \text{rank}(X)$, we compute the singular value decomposition

$$X = \sum_{i=1}^{\rho} \lambda_i \xi_i \eta_i^T;$$

here $(\xi_i)_{1 \leq i \leq \rho}$ is an orthonormal family in $\mathbf{R}^n$, $(\eta_i)_{1 \leq i \leq \rho}$ is an orthonormal family in $\mathbf{R}^K$, and we assume that the singular values $\lambda_i$ are in decreasing order (i.e., $\lambda_1 \geq \cdots \geq \lambda_\rho$). This can be accomplished in polynomial time within any numerical precision. Then $\widehat{X} = \sum_{i=1}^{\min\{k,\rho\}} \lambda_i \xi_i \eta_i^T$ is easily verified to be a rank $k$ approximation.

In addition to the operator norm, we are going to work with the *Frobenius norm*

$$\|X\|_F = \sqrt{\sum_{i=1}^{n} \sum_{j=1}^{K} x_{ij}^2}.$$

Although the following fact is well known, we provide its proof for completeness.

**Lemma 3.2.** *If $X$ has rank $k$, then $\|X\|_F^2 \leq k\|X\|^2$.*

*Proof.* Let $X = \sum_{i=1}^k \lambda_i \xi_i \eta_i^T$ be a singular value decomposition as above. Then

$$\|X\|_F^2 = \sum_{i,j} x_{ij}^2 = \sum_{i,j=1}^k \lambda_i \lambda_j \langle \xi_i, \xi_j \rangle \langle \eta_i, \eta_j \rangle.$$

Since $\xi_1, \ldots, \xi_k$ and $\eta_1, \ldots, \eta_k$ are orthonormal families, we have $\langle \xi_i, \xi_j \rangle = \langle \eta_i, \eta_j \rangle = 1$ if $i = j$ and $\langle \xi_i, \xi_j \rangle = \langle \eta_i, \eta_j \rangle = 0$ if $i \neq j$. Hence, $\|X\|_F^2 = \sum_{i=1}^k \lambda_i^2$. This implies the assertion, because $\lambda_i \leq \|X\|$ for all $1 \leq i \leq k$. □

### 3.2. Description of the algorithm

**Algo 3.3.** PARTITION$(A, k)$
*Input:* A $n \times K$ matrix $A$ and the parameter $k$. *Output:* A partition $S_1, \ldots, S_k$ of $V$.

1. Compute a rank $k$ approximation $\widehat{A}$ of $A$.
   For $j = 1, \ldots, 2\log K$ do
2. Let $\Gamma_j = K2^{-j}$ and compute $Q^{(j)}(v) = \{w \in V : \|\widehat{A}_w - \widehat{A}_v\|^2 \leq 0.01\Gamma_j^2\}$ for all $v \in V$.
   Then, determine sets $Q_1^{(j)}, \ldots, Q_k^{(j)}$ as follows: for $i = 1, \ldots, k$ do
3. Pick $v \in V \setminus \bigcup_{l=1}^{i-1} Q_l^{(j)}$ such that $|Q^{(j)}(v) \setminus \bigcup_{l=1}^{i-1} Q_l^{(j)}|$ is maximum.
   Set $Q_i^{(j)} = Q^{(j)}(v) \setminus \bigcup_{l=1}^{i-1} Q_l^{(j)}$ and $\xi_i^{(j)} = \frac{1}{|Q_i^{(j)}|} \sum_{w \in Q_i^{(j)}} \widehat{A}_w$.
4. Partition the entire set $V$ as follows: first, let $S_i^{(j)} = Q_i^{(j)}$ for all $1 \leq i \leq k$.
   Then, add each $v \in V \setminus \bigcup_{l=1}^k Q_l^{(j)}$ to a set $S_i^{(j)}$ such that $\|\widehat{A}_v - \xi_i^{(j)}\|$ is minimum.
   Set $r_j = \sum_{i=1}^k \sum_{v \in S_i^{(j)}} \|\widehat{A}_v - \xi_i^{(j)}\|^2$.
5. Let $J$ be such that $r^* = r_J$ is minimum. Return $S_1^{(J)}, \ldots, S_k^{(J)}$.

The basic idea behind PARTITION is to classify each individual $v \in V$ according to its row vector $\widehat{A}_v$ in the rank $k$ approximation $\widehat{A}$. That is, two individuals $v, w$ are deemed to belong to the same population iff $\|\widehat{A}_v - \widehat{A}_w\|^2 \leq 0.01\Gamma^2$. Hence, PARTITION tries to determine sets $S_1, \ldots, S_k$ such that for any two $v, w$ in the same set $S_j$ the distance $\|\widehat{A}_v - \widehat{A}_w\|$ is small. To justify this approach, we show that $\widehat{A}$ is "close" to the expectation $\mathbb{E}$ of $A$ in the following sense.

**Lemma 3.4.** *There is a constant $C > 0$ such that $\sum_{v \in V} \|\widehat{A}_v - \mathbb{E}_v\|^2 \leq Ck\lambda^2$ whp.*

*Proof.* Since both $\widehat{A}$ and $\mathbb{E}$ have rank at most $k$, and as therefore $\widehat{A} - \mathbb{E}$ has rank at most $2k$, Lemma 3.2 yields

$$\sum_{v \in V} \|\widehat{A}_v - \mathbb{E}_v\|^2 = \|\widehat{A} - \mathbb{E}\|_F^2 \leq 2k\|\widehat{A} - \mathbb{E}\|^2.$$

Furthermore, $\|\widehat{A} - \mathbb{E}\| \leq \|\widehat{A} - A\| + \|\mathbb{E} - A\| \leq 2\|\mathbb{E} - A\|$, because $\widehat{A}$ is a rank $k$ approximation of $A$. As Theorem 2.4 implies that $\|A - \mathbb{E}\|^2 \leq C\lambda^2/8$ for a certain constant $C > 0$, we thus obtain $\sum_{v \in V} \|\widehat{A}_v - \mathbb{E}_v\|^2 \leq 8k\|A - \mathbb{E}\|^2 \leq Ck\lambda^2$. □

Observe that Lemma 3.4 implies that for *most* $v$ we have $\|\widehat{A}_v - \mathbb{E}_v\|^2 \leq 10^{-6}\Gamma$, say. For letting $z = |\{v : \|\widehat{A}_v - \mathbb{E}_v\|^2 > 10^{-6}\Gamma\}|$, we get

$$10^{-6}\Gamma z \leq \sum_{v \in V} \|\widehat{A}_v - \mathbb{E}_v\|^2 \leq Ck\lambda^2,$$

whence $z \ll n_{\min}$ due to our assumption that $n_{\min}\Gamma \gg k\lambda^2$. Thus, most rows of $\widehat{A}$ are close to the corresponding rows of the *expected* matrix $\mathbb{E}$. Therefore, the separation assumption $n > \frac{C_k \sigma^2}{\gamma \omega}$ from Theorem 1.3 implies that for most pairs of elements in different classes $v \in V_i$, $w \in V_j$ the squared distance $\|\widehat{A}_v - \widehat{A}_w\|^2$ will be large (at least $0.99\Gamma$, say). By contrast, for most pairs $u, v \in V_i$ of elements belonging to the same class $\|\widehat{A}_v - \widehat{A}_w\|^2$ will be small (at most $0.01\Gamma$, say), because $\mathbb{E}_v = \mathbb{E}_w$.

As the above discussion indicates, if the algorithm were given $\Gamma$ as an input parameter, the procedure described in Steps 2–4 (with $\Gamma_j$ replaced by $\Gamma$) would yield the desired partition of $V$. The procedure described in Steps 2–4 is very similar to the spectral partitioning algorithm from McSherry (2001).

However, since $\Gamma$ is not given to the algorithm as an input parameter, PARTITION has to estimate $\Gamma$ on its own. (This is the new aspect here in comparison to McSherry (2001), and this fact necessitates a significantly more involved analysis.) To this end, the outer loop goes through $2 \log K$ "candidate values" $\Gamma_j$. These values are then used to obtain partitions $Q_1^{(j)}, \ldots, Q_k^{(j)}$ in Steps 2–4. More precisely, Step 2 uses $\Gamma_j$ to compute for each $v \in V$ the set $Q(v)$ of elements $w$ such that $\|\widehat{A}_w - \widehat{A}_v\| \leq 0.01\Gamma_j^2$. Then, Step 3 tries to compute "big" disjoint $Q_1^{(j)}, \ldots, Q_k^{(j)}$, where each $Q_i^{(j)}$ results from some $Q(v_i)$. Further, Step 4 assigns all elements $v$ not covered by $Q_1^{(j)}, \ldots, Q_k^{(j)}$ to that $Q_i^{(j)}$ whose "center vector" $\xi_i^{(j)}$ is closest to $\widehat{A}_v$. In addition, Step 4 computes an "error parameter" $r_j$. Finally, Step 5 outputs the partition that minimizes the error parameter $r_j$.

Thus, we need to show that eventually picking the partition whose error term $r_j$ is minimum yields a good approximation to the ideal partition $V_1, \ldots, V_k$. The basic reason why this is true is that the "empirical" mean $\xi_i^{(j)}$ should approximate the expectation $\mathbb{E}^{V_i}$ for class $V_i$ well iff $Q_i^{(j)}$ is a good approximation of $V_i$. Hence, if $Q_1^{(j)}, \ldots, Q_k^{(j)}$ is "close" to $V_1, \ldots, V_k$, then

$$r_j = \sum_{i=1}^{k} \sum_{v \in S_i^{(j)}} \|\widehat{A}_v - \xi_i^{(j)}\|^2$$

will be about as small as $\|\widehat{A} - \mathbb{E}\|_F^2$ (cf. Lemma 3.4). In fact, the following lemma shows that if $\Gamma_j$ is "close" to the ideal $\Gamma$, then $r_j$ will be small.

**Lemma 3.5.** *If $\frac{1}{2}\Gamma \leq \Gamma_j \leq \Gamma$, then $r_j \leq C_0 k^3 \lambda^2$ for a certain constant $C_0 > 0$.*

We defer the proof of Lemma 3.5 to Section 3.3. Furthermore, the next lemma shows that any partition such that $r_j$ is small yields a good approximation to $V_1, \ldots, V_k$.

**Lemma 3.6.** *Let $S_1, \ldots, S_k$ be a partition and $\xi_1, \ldots, \xi_k$ a sequence of vectors such that $\sum_{i=1}^{k} \sum_{v \in S_i} \|\xi_i - \widehat{A}_v\|^2 \leq C_0 k^3 \lambda^2$. Then there is a bijection $\Xi : \{1, \ldots, k\} \to \{1, \ldots, k\}$ such that the following holds.*

1. *$\|\xi_i - \mathbb{E}^{V_{\Xi(i)}}\|^2 \leq 0.001 \Gamma^2$ for all $i = 1, \ldots, k$, and*
2. *$\sum_{i=1}^{k} |S_i \triangle V_{\Xi(i)}| < 0.001 n_{\min}$.*

The proof of Lemma 3.6 can be found in Section 3.4.

*Proof of Claim 3.1.* Since the rank $k$ approximation $\widehat{A}$ can be computed in polynomial time (within any numerical precision), Partition is a polynomial time algorithm. Hence, we just need to show that $K^{-1} \leq \Gamma \leq K$; for then Partition will eventually try a $\Gamma_j$ such that $\frac{1}{2}\Gamma \leq \Gamma_j \leq \Gamma$, so that the claim follows from Lemmas 3.5 and 3.6. To see that $K^{-1} \leq \Gamma \leq K$, recall that we explicitly assume that $\Gamma > K^{-1}$. Furthermore, all entries of the vectors $\mathbb{E}^{V_i}$ lie between 0 and 1, whence $\Gamma = \max_{i<j} \|\mathbb{E}^{V_i} - \mathbb{E}^{V_j}\|^2 \leq K$. □

### 3.3. Proof of Lemma 3.5

Suppose that $\frac{1}{2}\Gamma \leq \Gamma_j \leq \Gamma$. To ease up the notation, we omit the superscript $j$; thus, we let $S_i = S_i^{(j)}$, $Q_i = Q_i^{(j)}$ for $1 \leq i \leq k$, and $Q(v) = Q^{(j)}(v)$ for $v \in V$ (cf. Steps 2–4 of Partition). The following lemma shows that there is a permutation $\pi$ such that $\xi_i$ is "close" to $\mathbb{E}^{V_{\pi(i)}}$ for all $1 \leq i \leq k$, and that the sets $Q_i$ are "not too small".

**Lemma 3.7.** *Suppose that $\frac{1}{2}\Gamma \leq \Gamma_j \leq \Gamma$. There is a bijection $\pi : \{1, \ldots, k\} \to \{1, \ldots, k\}$ such that for each $1 \leq i \leq k$ we have $|Q_i| \geq \frac{1}{2}|V_{\pi(i)}|$ and $\|\xi_i - \mathbb{E}^{V_{\pi(i)}}\|^2 \leq 0.1\Gamma$.*

*Proof.* For $1 \leq i \leq k$ we choose $\pi(i)$ so that $|Q_i \cap V_{\pi(i)}|$ is maximum. We shall prove below that for all $1 \leq l \leq k$ we have

$$\|\xi_l - \mathbb{E}^{V_{\pi(l)}}\|^2 \leq 0.1\Gamma, \tag{3.2}$$
$$|Q_l| \geq \max\{|V_i| : i \in \{1, \ldots, k\} \setminus \pi(\{1, \ldots, l-1\})\} - 0.01 n_{\min}, \tag{3.3}$$
$$|Q_l \cap V_{\pi(l)}| \geq |Q_l| - 0.01 n_{\min}. \tag{3.4}$$

These three inequalities imply the assertion. To see that $\pi$ is a bijection, let us assume that $\pi(l) = \pi(l')$ for two indices $1 \leq l < l' \leq k$. Indeed, suppose that $l = \min \pi^{-1}(l)$. Then $|Q_l| \geq |V_{\pi(l)}| - 0.01 n_{\min}$ by (3.3), and thus $|V_{\pi(l)} \setminus Q_l| \leq 0.1 n_{\min}$ by (3.4). As $Q_l \cap Q_{l'} = \emptyset$ by construction, we obtain the contradiction

$$0.99 n_{\min} \stackrel{(3.3)}{\leq} |Q_{l'}| \stackrel{(3.4)}{\leq} 1.1|Q_{l'} \cap V_{\pi(l)}| \leq 1.1|V_{\pi(l)} \setminus Q_l| \leq 0.11 n_{\min}.$$

Finally, as $\pi$ is bijective, (3.3) entails that $|Q_l| \geq 0.9 V_{\pi(l)}$ for all $1 \leq l \leq k$. Hence, due to (3.4) we obtain $|Q_l \cap V_l| \geq 0.9|Q_l| \geq \frac{1}{2}|V_{\pi(l)}|$, as desired.

The remaining task is to establish (3.2)–(3.4). We proceed by induction on $l$. Thus, let us assume that (3.2)–(3.4) hold for all $l < L$; we are to show that then (3.2)–(3.4) are true for $l = L$ as well. As a first step, we establish (3.3). To this end, consider a class $V_i$ such that $i \notin \pi(\{1, \ldots, L-1\})$ and let $Z_i = \{v \in V_i : \|\widehat{A}_v - \mathbb{E}_v\|^2 \leq 0.001\Gamma\}$. Then $0.001\Gamma(|V_i| - |Z_i|) \leq \sum_{v \in V_i \setminus Z_i} \|\widehat{A}_v - \mathbb{E}_v\|^2 \leq \|\widehat{A} - \mathbb{E}\|_F^2 \leq Ck\lambda^2$ (cf. Lemma 3.4) whence the assumption (3.1) on $\Gamma$ yields

$$|Z_i| \geq |V_i| - 0.01 n_{\min}, \qquad (3.5)$$

provided that $C_k$ is sufficiently large. Moreover, for all $v \in Z_i$ we have

$$Q(v) = \{w \in V : \|\widehat{A}_v - \widehat{A}_w\|^2 \leq 0.01\Gamma_j\} \supset Z_i, \qquad (3.6)$$

because we are assuming that $\Gamma_j \geq \Gamma/2$. In addition, let $w \in Q_l$ for some $l < L$; since our choice of $i$ ensures that $v \in V_i \neq V_{\pi(l)}$, we have

$$\sqrt{\Gamma} \leq \|\mathbb{E}^{V_{\pi(l)}} - \mathbb{E}_v\| \leq \|\mathbb{E}_v - \widehat{A}_v\| + \|\widehat{A}_w - \widehat{A}_v\| + \|\xi_l - \widehat{A}_w\| + \|\xi_l - \mathbb{E}^{V_{\pi(l)}}\|. \qquad (3.7)$$

Now, the construction in Step 3 of Partition ensures that $\|\widehat{A}_w - \xi_l\| \leq 0.1\sqrt{\Gamma}$. Furthermore, $\|\xi_l - \mathbb{E}^{V_{\pi(l)}}\| \leq \sqrt{\Gamma}/3$ by induction (cf. (3.2)), and $\|\widehat{A}_v - \mathbb{E}_v\| \leq 0.1\sqrt{\Gamma}$, because $v \in Z_i$. Hence, (3.7) entails that $\|\widehat{A}_w - \widehat{A}_v\| > 0.1\sqrt{\Gamma}$, so that $w \notin Q(v)$. Consequently, (3.6) yields

$$Z_i \cap Q_l = \emptyset \text{ for all } l < L. \qquad (3.8)$$

Finally, let $v_L$ signify the element chosen by Step 3 of Partition to construct $Q_L$. Then by construction $|Q_L| = |Q(v_L) \setminus \bigcup_{l=1}^{L-1} Q_l| \geq |Q(v) \setminus \bigcup_{l=1}^{L-1} Q_l|$. Therefore,

$$|Q_L| \geq |Q(v) \setminus \bigcup_{l=1}^{L-1} Q_l| \stackrel{(3.6),(3.8)}{\geq} |Z_i| \stackrel{(3.5)}{\geq} |V_i| - 0.01 n_{\min}.$$

As this estimate holds for all $i \notin \pi(\{1, \ldots, L-1\})$, (3.3) follows.

Thus, we know that $Q_L$ is "big". As a next step, we prove (3.4), i.e., we show that $Q_L$ "mainly" consists of vertices in $V_{\pi(L)}$. To this end, let $1 \leq i \leq k$ be such that $\|\mathbb{E}^{V_i} - \widehat{A}_{v_L}\|$ is minimum. Let $Y = Q_L \setminus V_i$. Then for all $w \in Y$ we have $\|\mathbb{E}_w - \widehat{A}_{v_L}\| \geq \|\mathbb{E}^{V_i} - \widehat{A}_v\|$. Further, since $\sqrt{\Gamma} \leq \|\mathbb{E}_w - \mathbb{E}^{V_i}\| \leq \|\mathbb{E}_w - \widehat{A}_{v_L}\| + \|\mathbb{E}^{V_i} - \widehat{A}_{v_L}\| \leq 2\|\mathbb{E}_w - \widehat{A}_{v_L}\|$, we conclude that $\|\mathbb{E}_w - \widehat{A}_{v_L}\|^2 \geq \frac{1}{4}\Gamma$. On the other hand, as $w \in Q_L$, we have $\|\widehat{A}_w - \widehat{A}_{v_L}\|^2 \leq 0.01\Gamma$. Therefore, we obtain $\|\widehat{A}_w - \mathbb{E}_w\|^2 \geq 0.1\Gamma$ for all $w \in Y$, so that

$$0.1|Y|\Gamma \leq \sum_{w \in Y} \|\widehat{A}_w - \mathbb{E}_w\|^2 \leq \|\widehat{A} - \mathbb{E}\|_F^2 \stackrel{\text{Lemma 3.4}}{\leq} c_k \lambda^2. \qquad (3.9)$$

Hence, due to our assumption (3.1) on $\Gamma$, (3.9) yields that $|Y| < 0.01 n_{\min}$. Consequently, (3.3) entails that $|V_i \cap Q_L| \geq 0.99|Q_L|$, so that $i = \pi(L)$. Hence, we obtain $|Q_L \cap V_{\pi(L)}| = |Q_L \cap V_i| = |Q_L \setminus Y| \geq |Q_L| - 0.01 n_{\min}$, thereby establishing (3.4).

Finally, to show (3.2), we note that by construction $\|\xi_L - \widehat{A}_{v_L}\|^2 \leq 0.01\Gamma$ and $\|\widehat{A}_w - \widehat{A}_{v_L}\|^2 \leq 0.01\Gamma$ for all $w \in Q_L \cap V_{\pi(L)}$ (cf. Step 3 of `Partition`). Therefore,

$$
\begin{aligned}
|Q_L \cap V_{\pi(L)}| \cdot \|\mathbb{E}_{\pi(L)} - \xi_L\|^2 \\
\leq \quad & 3 \sum_{w \in Q_L \cap V_{\pi(L)}} \|\xi_L - \widehat{A}_{v_L}\|^2 + \|\widehat{A}_w - \widehat{A}_{v_L}\|^2 + \|\widehat{A}_w - \widehat{\mathbb{E}}_{\pi(L)}\|^2 \\
\leq \quad & 0.06\Gamma |Q_L \cap V_{\pi(L)}| + 3\|\widehat{A} - \mathbb{E}\|_F^2 \\
\overset{\text{Lemma 3.4}}{\leq} \quad & 0.06\Gamma |Q_L \cap V_{\pi(L)}| + 3c_k \lambda^2. \quad (3.10)
\end{aligned}
$$

Since $|Q_L \cap V_{\pi(L)}| \geq 0.9 n_{\min}$ due to (3.3) and (3.4), (3.10) entails that $\|\mathbb{E}_{\pi(L)} - \xi_L\|^2 \leq 0.07\Gamma + \frac{6c_k \lambda^2}{n_{\min}} \leq 0.1\Gamma$. Thus, (3.2) follows. □

In the sequel, we shall assume without loss of generality that the map $\pi$ from Lemma 3.7 is just the identity, i.e., $\pi(i) = i$ for all $i$. Bootstrapping on the estimate $\|\xi_i - \mathbb{E}^{V_i}\|^2 \leq 0.1\Gamma$ for $1 \leq i \leq k$ from Lemma 3.7, we derive the following stronger estimate.

**Corollary 3.8.** *For all $1 \leq i \leq k$ we have $\|\xi_i - \mathbb{E}^{V_i}\|^2 \leq 100|Q_i|^{-1} \sum_{v \in Q_i} \|\widehat{A}_v - \mathbb{E}_v\|^2$.*

*Proof.* By the Cauchy-Schwarz inequality,

$$\|\xi_i - \mathbb{E}^{V_i}\| = |Q_i|^{-1} \left\| \sum_{v \in Q_i} \widehat{A}_v - \mathbb{E}^{V_i} \right\| \leq |Q_i|^{-1/2} \left[ \sum_{v \in Q_i} \|\widehat{A}_v - \mathbb{E}^{V_i}\|^2 \right]^{1/2}. \quad (3.11)$$

Furthermore, as $\|\xi_i - \mathbb{E}^{V_i}\|^2 \leq 0.1\Gamma$ by Lemma 3.7, for all $v \in Q_i \setminus V_i$ we have

$$\|\widehat{A}_v - \mathbb{E}^{V_i}\|^2 \leq 2(\|\widehat{A}_v - \xi_i\|^2 + \|\xi_i - \mathbb{E}^{V_i}\|^2) \leq \Gamma^{1/3}, \quad (3.12)$$

because the construction of $Q_i$ in Step 3 of `Partition` ensures that $\|\widehat{A}_v - \xi_i\|^2 \leq 0.01\Gamma$. Hence, as $\|\mathbb{E}_v - \mathbb{E}^{V_i}\|^2 \geq \Gamma$, (3.12) implies that $\|\widehat{A}_v - \mathbb{E}_v\| \geq 0.1\|\widehat{A}_v - \mathbb{E}^{V_i}\|$. Therefore, the assertion follows from (3.11). □

**Corollary 3.9.** *For all $v \in S_i \setminus V_i$ we have $\|\widehat{A}_v - \xi_i\| \leq 3\|\widehat{A}_v - \mathbb{E}_v\|$.*

*Proof.* Let $i \neq l$ and consider a $v \in S_i \cap V_l$. We shall establish below that

$$\|\widehat{A}_v - \xi_i\| \leq \|\widehat{A}_v - \xi_l\|. \quad (3.13)$$

Then by Lemma 3.7 $\|\widehat{A}_v - \xi_i\| \leq \|\widehat{A}_v - \mathbb{E}_v\| + \|\mathbb{E}_v - \xi_l\| \leq \|\widehat{A}_v - \mathbb{E}_v\| + \sqrt{\Gamma}/3$, and thus $\sqrt{\Gamma} \leq \|\mathbb{E}_v - \mathbb{E}^{V_i}\| \leq \|\widehat{A}_v - \xi_i\| + \|\xi_i - \mathbb{E}^{V_i}\| + \|\widehat{A}_v - \mathbb{E}_v\| \leq 2\|\widehat{A}_v - \mathbb{E}_v\| + \frac{2}{3}\sqrt{\Gamma}$.

Consequently, we obtain $\|\widehat{A}_v - \mathbb{E}_v\| \geq \frac{1}{6}\sqrt{\Gamma}$, so that the assertion follows from the estimate

$$\|\widehat{A}_v - \xi_i\| \stackrel{(3.13)}{\leq} \|\widehat{A}_v - \xi_l\| \leq \|\widehat{A}_v - \mathbb{E}_v\| + \|\mathbb{E}_v - \xi_l\|$$
$$\stackrel{\text{Lemma 3.7}}{\leq} \|\widehat{A}_v - \mathbb{E}_v\| + \frac{\sqrt{\Gamma}}{3} \leq 3\|\widehat{A}_v - \mathbb{E}_v\|.$$

Finally, we prove (3.13). If $v \in S_i \cap V_l \setminus Q_i$, then the construction of $S_i$ in Step 4 of `Partition` guarantees that $\|\widehat{A}_v - \xi_i\| \leq \|\widehat{A}_v - \xi_l\|$, as claimed. Thus, assume that $v \in Q_i \cap V_l$. Then

$$\|\widehat{A}_v - \xi_i\| \leq 0.15\sqrt{\Gamma} \qquad \text{[by the definition of } Q_i \text{ in Step 3]},$$
$$\max\{\|\xi_i - \mathbb{E}^{V_i}\|, \|\xi_l - \mathbb{E}_v\|\} \leq \frac{1}{3}\sqrt{\Gamma} \qquad \text{[by Lemma 3.7]},$$
$$\|\mathbb{E}^{V_i} - \mathbb{E}_v\| \geq \sqrt{\Gamma} \qquad .$$

Therefore, if $\|\widehat{A}_v - \xi_l\| < \|\widehat{A}_v - \xi_i\|$, then we would arrive at the contradiction

$$\sqrt{\Gamma} \leq \|\mathbb{E}^{V_i} - \mathbb{E}_v\| \leq \|\mathbb{E}^{V_i} - \xi_i\| + \|\mathbb{E}_v - \xi_l\| + \|\xi_i - \xi_l\|$$
$$\leq \frac{2}{3}\sqrt{\Gamma} + \|\widehat{A}_v - \xi_i\| + \|\widehat{A}_v - \xi_l\| < \frac{2}{3}\sqrt{\Gamma} + 2\|\widehat{A}_v - \xi_i\| \leq 0.99\sqrt{\Gamma}.$$

Thus, we conclude that $\|\widehat{A}_v - \xi_l\| \geq \|\widehat{A}_v - \xi_i\|$, thereby completing the proof. □

*Proof of Lemma 3.5.* Since $|Q_i| \geq \frac{1}{2}|V_i|$ by Lemma 3.7, we have the estimate

$$\sum_{i=1}^k \sum_{w \in S_i \cap V_i} \|\widehat{A}_w - \xi_i\|^2 \leq 2\sum_{i=1}^k \sum_{w \in S_i \cap V_i} \left[\|\widehat{A}_w - \mathbb{E}_w\|^2 + \|\mathbb{E}_w - \xi_i\|^2\right]$$
$$\stackrel{\text{Cor. 3.8}}{\leq} 2\|\widehat{A} - \mathbb{E}\|_F^2 + 200\sum_{i=1}^k \frac{|S_i \cap V_i|}{|Q_i|} \sum_{v \in Q_i} \|\widehat{A}_v - \mathbb{E}_v\|^2$$
$$\leq 500\|\widehat{A} - \mathbb{E}\|_F^2. \qquad (3.14)$$

Furthermore, by Corollary 3.9

$$\sum_{i=1}^k \sum_{v \in S_i \setminus V_i} \|\widehat{A}_v - \xi_i\|^2 \leq 9\sum_{i=1}^k \sum_{v \in S_i \setminus V_i} \|\widehat{A}_v - \mathbb{E}_v\|^2 \leq 9\|\widehat{A} - \mathbb{E}\|_F^2. \qquad (3.15)$$

Since $\|\widehat{A} - \mathbb{E}\|_F^2 \leq Ck\lambda^2$ by Lemma 3.4, the bounds (3.14) and (3.15) imply the assertion. □

### 3.4. Proof of Lemma 3.6

Set $S_{ab} = S_a \cap V_b$ for $1 \leq a, b \leq k$. Moreover, for each $1 \leq a \leq k$ let $1 \leq \pi(a) \leq k$ be such that $\|\mathbb{E}^{V_{\pi(a)}} - \xi_a\|$ is minimum. Then for all $b \neq \pi(a)$ we have

$$\sqrt{\Gamma} \leq \|\mathbb{E}^{V_{\pi(a)}} - \mathbb{E}^{V_b}\| \leq \|\mathbb{E}^{V_{\pi(a)}} - \xi_a\| + \|\mathbb{E}^{V_b} - \xi_a\| \leq 2\|\mathbb{E}^{V_b} - \xi_a\|, \qquad (3.16)$$

so that $\|\mathbb{E}^{V_b} - \xi_a\| \geq \sqrt{\Gamma}/2$. Therefore, by our assumption that $\sum_{i=1}^{k} \sum_{v \in S_i} \|\xi_i - \widehat{A}_v\|^2 \leq C_0 k^3 \lambda^2$, we have

$$\begin{aligned}
\frac{\Gamma}{4} \sum_{a=1}^{k} \sum_{1 \leq b \leq k: b \neq \pi(a)} |S_{ab}| &\leq \sum_{a,b=1}^{k} |S_{ab}| \cdot \|\mathbb{E}^{V_b} - \xi_a\|^2 \\
&\leq 2 \sum_{a,b=1}^{k} \sum_{v \in S_{ab}} \|\mathbb{E}_v - \widehat{A}_v\|^2 + \|\widehat{A}_v - \xi_a\|^2 \\
&\leq 2\|\widehat{A} - \mathbb{E}\|_F^2 + 2 \sum_{a,b=1}^{k} \sum_{v \in S_{ab}} \|\widehat{A}_v - \xi_a\|^2 \\
&\stackrel{\text{Lemma 3.4}}{\leq} 4C_0 k^3 \lambda^2 + 2C_0 k^3 \lambda^2 \leq C_0^2 k^3 \lambda^2. \quad (3.17)
\end{aligned}$$

Hence,

$$\sum_{a=1}^{k} |S_a \triangle V_{\pi(a)}| = \sum_{1 \leq a,b \leq k: b \neq \pi(a)} 2|S_{ab}| \leq \frac{8c_0^2 k^3 \lambda^2}{\Gamma} \leq 0.001 n_{\min}, \quad (3.18)$$

provided that $C_k$ is sufficiently large (cf. (3.1)). Combining (3.17) and (3.18), we obtain $\frac{n_{\min}}{2} \|\mathbb{E}^{V_{\pi(a)}} - \xi_a\|^2 \leq |S_a \cap V_{\pi(a)}| \cdot \|\mathbb{E}_{\pi(a)} - \xi_a\|^2 \leq c_0^2 k^3 \lambda^2$, whence

$$\|\mathbb{E}_{\pi(a)} - \xi_a\|^2 \leq \frac{2c_0^2 k^3 \lambda^2}{n_{\min}} \leq 0.001\Gamma \quad \text{for all } 1 \leq a \leq k, \quad (3.19)$$

provided that $C_k$ is large enough. Thus, we have established the first two parts of the lemma. In addition, observe that (3.18) implies that $\pi$ is bijective (because the sets $S_1, \ldots, S_k$ are pairwise disjoint and $|V_a| \geq n_{\min}$ for all $1 \leq a \leq k$). Finally, the third assertion follows from the estimate

$$\begin{aligned}
\sum_{a,b=1}^{k} |S_{ab}| \cdot \|\mathbb{E}^{V_{\pi(a)}} - \mathbb{E}^{V_{\pi(b)}}\|^2 &\leq 2 \sum_{a,b=1}^{k} |S_{ab}| \left(\|\mathbb{E}^{V_{\pi(a)}} - \xi_a\|^2 + \|\mathbb{E}^{V_{\pi(b)}} - \xi_a\|^2\right) \\
&\stackrel{(3.16)}{\leq} 8 \sum_{a,b=1}^{k} |S_{ab}| \cdot \|\mathbb{E}^{V_{\pi(b)}} - \xi_a\|^2 \\
&\stackrel{(3.17)}{\leq} 8 C_0^2 k^3 \lambda^2 \leq 0.001 \Gamma n_{\min},
\end{aligned}$$

where we assume once more that $C_k$ is sufficiently large.

## 4. Experiments

We illustrate the effectiveness of spectral techniques using simulations. In particular, we explore the case when we have a mixture of two populations; we show that when $NK > 1/\gamma^2$ and $K > 1/\gamma$, either the first or the second left singular vector of $X$ shows an approximately correct partitioning, meaning that the success rate is well above $1/2$. The entry-wise expected value matrix $\mathcal{X}$ is: among $K/2$ features, $p_1^i > p_2^i$ and for the other half, $p_1^i < p_2^i$ such that $\forall i$, $p_1^i, p_2^i \in \{\frac{1+\alpha}{2} + \frac{\epsilon}{2}, \frac{1-\alpha}{2} + \frac{\epsilon}{2}\}$, where $\epsilon = 0.1\alpha$. Hence $\gamma = \alpha^2$. We report results on balanced cases only, but we do observe that unbalanced cases show similar tradeoffs. For each population $P$, the success rate is defined as the number of individuals that are correctly classified, i.e., they belong to a group that $P$ is the majority of that group, versus the size of the population $|P|$.

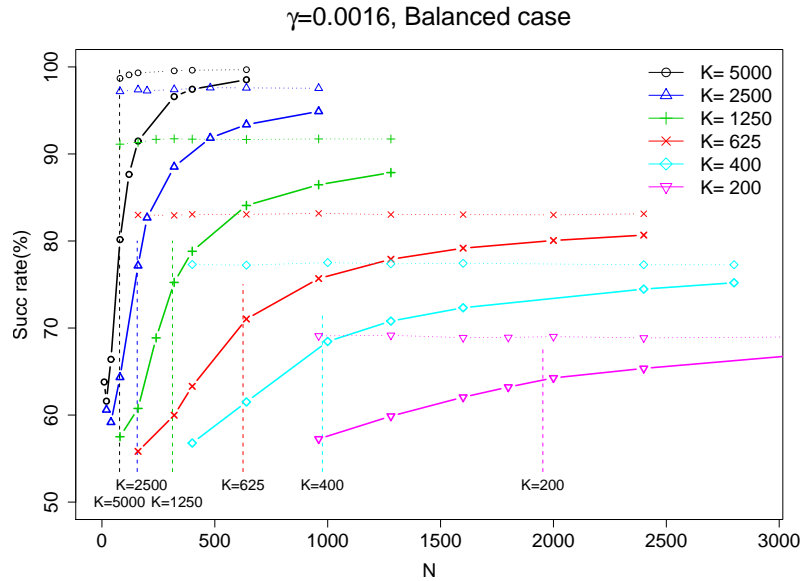

FIG 1. *Plots show success rate as a function of $N$ for several values of $K$, when $\gamma = (0.04)^2$. Each point is an average over 100 trials. Horizontal lines ("oracles") indicate the information-theoretically best possible success rate for that value of $K$ (how well one could do if one knew in advance which features satisfied $p_1^i > p_2^i$ and which satisfied $p_1^i < p_2^i$; they are not exactly horizontal because they are also an average over 100 runs). Vertical bars indicate the value of $N$ for which $NK = 1/\gamma^2$.*

Each point on the SVD curve corresponds to an average rate over 100 trials. Since we are interested in exploring the tradeoffs of $N, K$ in all ranges (e.g., when $N << K$ or $N >> K$), rather than using the threshold $T$ in Procedure Classify that is chosen in case both $N, K > 1/\gamma$, to decide which singular vector to use, we try both $u_1$ and $u_2$ and use the more effective one to measure the success rate at each trial. For each data point, the distribution of $X$ is fixed across all trials and we generate an independent $X_{2N \times K}$ for each trial to measure success rate based on the more effective classifier between $u_1$ and $u_2$.

One can see from the plot that when $K < 1/\gamma$, i.e., when $K = 200$ and 400, no matter how much we increase $N$, the success rate is consistently low. Note that $50/100$ of success rate is equivalent to a total failure. In contrast, when $N$ is smaller than $1/\gamma$, as we increase $K$, we can always classify with a high success rate, where in general, $NK > 1/\gamma^2$ is indeed necessary to see a high success rate. In particular, the curves for $K = 5000, 2500, 1250$ show the sharpness of the threshold behavior for increasing sample size $n$ from below $1/K\gamma^2$ to above. For each curve, we also compute the best possible classification one could hope to make if one knew in advance which features satisfied $p_1^i > p_2^i$ and which satisfied $p_1^i < p_2^i$. These are the horizontal(ish) dotted lines above each curve. The fact that the solid curves are approaching these information-theoretic upper bounds shows that the spectral technique is correctly using the available information.

## Acknowledgements

We thank John Lafferty, Frank McSherry, Roman Vershynin and Larry Wasserman for many helpful discussions on this work.

## Appendix A: More Proofs for the simple algorithm classify

### A.1. Proof of Lemma 2.1

Let $u_1, \ldots, u_n, v_1, \ldots, v_n$ be the $n$ left and right singular vectors of $X$, corresponding to $s_1(X) \geq s_2(X) \geq \cdots \geq s_n(X)$, we have for $\forall i, \|u_i\|_2 = 1, \|v_i\|_2 = 1$ such that $X^T u_i = s_i(X) v_i$ and $X v_i = s_i(X) u_i$.

Before we prove Lemma 2.1, given an $n \times K$ matrix $X$, where $n < K$, let us first define $H = XX^T$ and a block matrix

$$Y = \begin{bmatrix} 0 & X \\ X^T & 0 \end{bmatrix}_{(2N+K) \times (2N+K)}. \tag{A.1}$$

Recall that singular values of a real $n \times K$ matrix $X$ are exactly the non-negative square roots of the $n$ largest eigenvalues of $H = XX^T$, i.e, $s_i(X) = \sqrt{\lambda_i(H)}, \forall i = 1, \ldots, n$, given that

$$H u_i = XX^T u_i = s_i(X) X v_i = s_i^2(X) u_i. \tag{A.2}$$

Hence the left singular vectors $u_1, \ldots, u_n$ of $X$ are eigenvectors of $H$ corresponding to $\lambda_i(H) = s_i^2(X)$.

We next show that the first $n$ eigenvalues of $Y$ and their corresponding eigenvectors:

$$Y \begin{bmatrix} u_i \\ v_i \end{bmatrix} = \begin{bmatrix} 0 & X \\ X^T & 0 \end{bmatrix} \begin{bmatrix} u_i \\ v_i \end{bmatrix} = \begin{bmatrix} X v_i \\ X^T u_i \end{bmatrix} = \begin{bmatrix} s_i(X) u_i \\ s_i(X) v_i \end{bmatrix} = s_i(X) \begin{bmatrix} u_i \\ v_i \end{bmatrix}, \tag{A.3}$$

and hence

**Proposition A.1.** *The largest $n$ eigenvalues of $Y$ are $s_1(X), \ldots, s_n(X)$ with corresponding eigenvectors $[u_i, v_i], \forall i = 1, \ldots, n$, where $u_i, v_i, \forall i$, are left and right singular vectors of $X$ corresponding to $s_i(X)$.*

In fact both $\pm s_i(X)$ are eigenvalues of $Y$, which is irrelevant.

*Proof of Lemma 2.1.* We first state a theorem, whose statement appears in a lecture note by Spielman (2002), with a slight modification (off by a factor on RHS). Our proof for this theorem is included here for completeness. It is known that for any real symmetric matrix, there exist a set of $n$ orthonormal eigenvectors.

**Theorem A.2** (Modified Version of Spielman (2002)). *For A and M being two symmetric matrices and $E = M - A$. Let $\lambda_1(A) \geq \lambda_2(A) \geq \cdots \geq \lambda_n(A)$ be eigenvalues of A, with orthonormal eigenvectors $v_1, v_2, \ldots, v_n$ and let $\lambda_1(M) \geq \lambda_2(M) \geq \cdots \geq \lambda_n(M)$ be eigenvalues of M and $w_1, w_2, \ldots, w_n$ be the corresponding orthonormal eigenvectors of M, with $\theta_i = \angle(v_i, w_i)$. Then*

$$\theta_i \sim \sin(\theta_i) \leq \frac{\|E - \Delta_i I\|_2}{\mathsf{gap}(i, A)} \leq \frac{\|E\|_2 + |\Delta_i|}{\mathsf{gap}(i, A)} \leq \frac{2\|E\|_2}{\mathsf{gap}(i, A)} \tag{A.4}$$

*where $\mathsf{gap}(i, A) = \min_{j \neq i} |\lambda_i(A) - \lambda_j(A)|$ and $\Delta_i = \lambda_i(M) - \lambda_i(A)$.*

Let us apply Theorem A.2 to the symmetric matrix $Y$ in (A.1). In particular, we only compare the first $n$ eigenvectors of $Y$ of $\mathcal{Y}$. For the numerator of RHS of (A.4), we have $E = Y - \mathcal{Y}$, and $\|E\|_2 = \|Y - \mathcal{Y}\|_2 = s_1(Y - \mathcal{Y})$ by a derivation similar to (A.3), where eigenvectors of $E$ are concatenations of left and right singular vectors of $X - \mathcal{X}$; For the denominator, we have by Proposition A.1, $\mathsf{gap}(i, \mathcal{Y}) = \min_{j \neq i} |\lambda_i(\mathcal{Y}) - \lambda_j(\mathcal{Y})| = \min_{j \neq i} |s_i(\mathcal{X}) - s_j(\mathcal{X})|$. □

We first prove the following claim.

**Claim A.3.** *For any symmetric $n \times n$ matrix A, let $\lambda_i, \forall i = 1, \ldots, n$ be eigenvalues of A with orthonormal eigenvectors $v_1, v_2, \ldots, v_n$, for all $y \perp v_i$,*

$$\|(A - \lambda_i)y\|_2 \geq \min_{j \neq i} |\lambda_i - \lambda_j| \|y\|_2.$$

*Proof.* Let us first assume $y \perp v_i$ and write $y = \sum_{j=1, j \neq i}^n c_j v_j$, thus we have $\|y\|_2 = \sqrt{\sum_{j=1, j \neq i}^n c_j^2}$ and

$$\begin{aligned}
\|(A - \lambda_i)y\|_2 &= \left\| \sum_{j=1, j \neq i}^n c_j (A - \lambda_i) v_j \right\|_2 \\
&= \left\| \sum_{j=1, j \neq i}^n c_j (\lambda_j - \lambda_i) v_j \right\|_2 = \sqrt{\sum_{j=1, j \neq i}^n c_j^2 |\lambda_j - \lambda_i|^2} \\
&\geq \min_{j \neq i} |\lambda_i - \lambda_j| \sqrt{\sum_{j=1, j \neq i}^n c_j^2} = \min_{j \neq i} |\lambda_i - \lambda_j| \|y\|_2.
\end{aligned}$$

□

*Proof of Theorem A.2.* Let us construct a vector $y$ that is orthogonal to $v_i$ as follows:

$$y = w_i - (v_i^T w_i) v_i$$

By Claim A.3, we have

$$\|(A - \lambda_i(A))y\|_2 \geq \min_{j \neq i} |\lambda_i(A) - \lambda_j(A)| \, \|y\|_2 \, ,$$

and hence

$$\|y\|_2 \leq \frac{\|(A - \lambda_i(A))y\|_2}{\min_{j \neq i} |\lambda_i(A) - \lambda_j(A)|}$$

On the other hand,

$$\begin{aligned}
\|(A - \lambda_i(A))y\|_2 &= \left\|(A - \lambda_i(A))(w_i - (v_i^T w_i)v_i)\right\|_2 \\
&= \|(A - \lambda_i(A))w_i\|_2 \\
&= \|(M - E - \lambda_i(A))w_i\|_2 \\
&= \|(\lambda_i(M) - \lambda_i(A))w_i - E w_i\|_2 \\
&= \|(\Delta_i I - E)w_i\|_2 \leq \|E - \Delta_i I\|_2 \\
&\leq \|E\|_2 + |\Delta_i|
\end{aligned}$$

Finally, given that $\|w\|_2 = 1$,

$$\begin{aligned}
\sin(\theta_i) &= \frac{\|y\|_2}{\|w\|_2} \leq \frac{\|(A - \lambda_i(A))y\|_2}{\min_{j \neq i} |\lambda_i(A) - \lambda_j(A)|} \\
&\leq \frac{\|E\|_2 + |\Delta_i|}{\mathsf{gap}(i, A)}.
\end{aligned}$$

**Lemma A.4.** $\forall i = 1, \ldots, n, \ |\Delta_i| \leq \|E\|_2$.

*Proof.* Let $S_j$ be a subspace of dimension $j$. Recall the following definition of $\lambda_i$ for a matrix:

$$\lambda_i(M) = \inf_{S_{N-i+1}} \sup_{x \in S_{N-i+1}, \|x\|_2 = 1} x^T M x. \tag{A.5}$$

In the following, let $S_{N-i+1}^v$ be the subspace that is orthogonal to the subset of orthonormal eigenvectors $v_1, \ldots, v_{i-1}$ of symmetric matrix $A$. Note that this is the $N - i + 1$ dimensional subspace that achieves the minimum of the maximum of $v^T A v$ over all unit-length vectors $v$ in the particular subspace. We have

$$\begin{aligned}
\lambda_i(M) &= \inf_{S_{N-i+1}} \sup_{x \in S_{N-i+1}, \|x\|_2 = 1} x^T M x \leq \sup_{x \in S_{N-i+1}^v, \|x\|_2 = 1} x^T M x \\
&\leq \sup_{x \in S_{N-i+1}^v, \|x\|_2 = 1} x^T (A + E) x \\
&\leq \sup_{v \in S_{N-i+1}^v, \|v\|_2 = 1} v^T A v + \sup_{x \in R^n, \|x\|_2 = 1} |x^T E x| \\
&= \lambda_i(A) + \|E\|_2 \, .
\end{aligned}$$

For the other direction, let $S^w_{N-i+1}$ be the subspace that is orthogonal to the subset of orthonormal eigenvectors $w_1, \ldots, w_{i-1}$ of symmetric matrix $M$. Note that this is the $N-i+1$ dimensional subspace that achieves the minimum of the maximum of $w^T M w$ over all unit-length vectors $w$ in the particular subspace. We have

$$
\begin{aligned}
\lambda_i(A) &= \inf_{S_{N-i+1}} \sup_{x \in S_{N-i+1}, \|x\|_2=1} x^T A x \leq \sup_{x \in S^w_{N-i+1}, \|x\|_2=1} x^T A x \\
&\leq \sup_{x \in S^w_{N-i+1}, \|x\|_2=1} x^T (M + (-E)) x \\
&\leq \sup_{w \in S^w_{N-i+1}, \|w\|_2=1} w^T M w + \sup_{x \in R^n, \|x\|_2=1} x^T (-E) x \\
&\leq \sup_{w \in S^w_{N-i+1}, \|w\|_2=1} w^T M w + \sup_{x \in R^n, \|x\|_2=1} |x^T (-E) x| \\
&= \lambda_i(M) + \|E\|_2,
\end{aligned}
$$

where $\|E\|_2 = \|-E\|_2$. Thus $-\|E\|_2 \leq \lambda_i(M) - \lambda_i(A) \leq \|E\|_2$, and $|\Delta_i| \leq \|E\|_2$. □

Therefore, $\sin(\theta_i) \leq \frac{\|E\|_2 + |\Delta_i|}{\mathsf{gap}(i,A)} \leq \frac{2\|E\|_2}{\mathsf{gap}(i,A)}$. □

### A.2. Some Propositions regarding the static matrices

For static matrix $\mathcal{H} = \mathcal{X}\mathcal{X}^T$ and $\mathcal{Y} = \begin{bmatrix} 0 & \mathcal{X} \\ \mathcal{X}^T & 0 \end{bmatrix}$, we define

$$
\begin{aligned}
\mathsf{gap}(\mathcal{H}) &= |\lambda_1(\mathcal{H}) - \lambda_2(\mathcal{H})|, \\
\mathsf{gap}(\mathcal{Y}) &= |\lambda_1(\mathcal{Y}) - \lambda_2(\mathcal{Y})| = \frac{\mathsf{gap}(\mathcal{H})}{\lambda_1(\mathcal{Y}) + \lambda_2(\mathcal{Y})},
\end{aligned}
$$

**Proposition A.5.** *For static matrix $\mathcal{Y}$, let $\mathsf{gap}(\mathcal{Y}) = |\lambda_1(\mathcal{Y}) - \lambda_2(\mathcal{Y})| = \frac{\mathsf{gap}(\mathcal{H})}{\lambda_1(\mathcal{Y}) + \lambda_2(\mathcal{Y})}$, we have*

$$
\begin{aligned}
\sqrt{\max\{N_1 a, N_2 c\}} &\leq \lambda_1(\mathcal{Y}) \leq \sqrt{N_1 a + N_2 c}, \\
\sqrt{N_1 a + N_2 c} &\leq \lambda_1(\mathcal{Y}) + \lambda_2(\mathcal{Y}) \leq \sqrt{2(N_1 a + N_2 c)}, \\
\sqrt{\frac{N_1 N_2 (ac - b^2)}{N_1 a + N_2 c}} &\leq \lambda_2(\mathcal{Y}) \leq \sqrt{\frac{2 N_1 N_2 (ac - b^2)}{N_1 a + N_2 c}}, \\
\mathsf{gap}(\mathcal{Y}) = \Theta\left(\frac{\mathsf{gap}(\mathcal{H})}{\sqrt{N_1 a + N_2 c}}\right) &= \Theta\left(\frac{\sqrt{(N_1 a + N_2 c)^2 - 4 N_1 N_2 (ac - b^2)}}{\sqrt{N_1 a + N_2 c}}\right).
\end{aligned}
$$

*Proof.* We first show the following:

*A. Blum et al./Separating populations with wide data: A spectral analysis*   105**Proposition A.6.** *For static matrix $\mathcal{H} = \mathcal{X}\mathcal{X}^T$ as in (2.12), Let $\lambda_1(\mathcal{H}), \lambda_2(\mathcal{H})$ be the non-zero eigenvalues of $\mathcal{H}$, and denote $\mathsf{gap}(\mathcal{H}) = |\lambda_1(\mathcal{H}) - \lambda_2(\mathcal{H})|$.*

$$\lambda_1(\mathcal{H}) = \frac{N_1 a + N_2 c + \sqrt{(N_1 a - N_2 c)^2 + 4 N_1 N_2 b^2}}{2}, \quad (A.6)$$

$$\lambda_2(\mathcal{H}) = \frac{N_1 a + N_2 c - \sqrt{(N_1 a - N_2 c)^2 + 4 N_1 N_2 b^2}}{2}, \quad (A.7)$$

$$|N_1 a - N_2 c| \leq \mathsf{gap}(\mathcal{H}) \leq N_1 a + N_2 c, \quad (A.8)$$

*where $\lambda_2(\mathcal{H}) = 0$, when $ac = b^2$ and $\mathsf{gap}(\mathcal{H}) = N_1 a + N_2 c$.*

*Proof.* Let $\mathcal{H} = \mathcal{X}\mathcal{X}^T$. The rank of $\mathcal{H}$ is at most 2. Therefore there exist at most two non-zero eigenvalues $\lambda_1, \lambda_2$ for $\mathcal{H}$, with corresponding nonzero eigenvectors $v_1, v_2$ being constant on each population. This is true because if we multiply $\mathcal{H} = \mathcal{X}\mathcal{X}^T$ by a permutation matrix $P$ to exchange two rows among the same population, we have $P\mathcal{H}v_i = \lambda_i P v_i, \forall i = 1, 2$; given that $P\mathcal{H}v_i = \mathcal{H}v_i$, we deduce that $Pv_i = v_i$ for non-zero $\lambda_i$. Hence $v_i$ must be constant on each population.

Let the top two eigenvector $v_1, v_2$ be of form $[x, \ldots, x, y, \ldots, y]$, where $x$ repeats $N_1$ times and $y$ repeats $N_2$ times; Note that they corresponds to $\bar{u}_1$ and $\bar{u}_2$ of $\mathcal{X}$ following a derivation similar to (A.2).

We thus have the following equations:

$$N_1 a x + N_2 b y = \lambda x, \quad (A.9)$$
$$N_1 b x + N_2 c y = \lambda y, \quad (A.10)$$

which can be written in a matrix form:

$$\begin{bmatrix} N_1 a - \lambda & N_2 b \\ N_1 b & N_2 c - \lambda \end{bmatrix} \begin{bmatrix} x \\ y \end{bmatrix} = 0$$

Given that

$$\begin{bmatrix} x \\ y \end{bmatrix} \neq 0,$$

the matrix is not one-to-one and therefore

$$D \begin{bmatrix} N_1 a - \lambda & N_2 b \\ N_1 b & N_2 c - \lambda \end{bmatrix} = 0.$$

By solving $(N_1 a - \lambda)(N_2 c - \lambda) - N_1 N_2 b^2 = 0$, we get $\lambda_1(\mathcal{H}), \lambda_2(\mathcal{H})$ and $\mathsf{gap}(\mathcal{H})$. We next derive an upper bound on $\mathsf{gap}(\mathcal{H})$.

$$\mathsf{gap}(\mathcal{H}) = \sqrt{(N_1 a - N_2 c)^2 + 4 N_1 N_2 b^2} \quad (A.11)$$
$$= \sqrt{(N_1 a + N_2 c)^2 - 4 N_1 N_2 a c + 4 N_1 N_2 b^2} \quad (A.12)$$
$$\leq \sqrt{(N_1 a + N_2 c)^2} \quad (A.13)$$
$$\leq N_1 a + N_2 c, \quad (A.14)$$

where $a, c \geq 0$ and $ac \geq b^2$ as in Proposition 2.10. Hence $\mathsf{gap}(\mathcal{H}) \geq |N_1 a - N_2 c|$, given that $b^2 \geq 0$. □

Thus we have

$$\max\{N_1 a, N_2 c\} \leq \lambda_1(\mathcal{H}) \leq N_1 a + N_2 c, \qquad (A.15)$$
$$0 \leq \lambda_2(\mathcal{H}) \leq \min\{N_1 a, N_2 c\}, \qquad (A.16)$$
$$\lambda_1(\mathcal{H}) + \lambda_2(\mathcal{H}) = N_1 a + N_2 c, \qquad (A.17)$$
$$\lambda_1(\mathcal{H})\lambda_2(\mathcal{H}) = N_1 N_2 (ac - b^2), \qquad (A.18)$$

Given two largest eigenvalues of $\mathcal{Y}$, $\lambda_1(\mathcal{Y}) = \sqrt{\lambda_1(\mathcal{H})}$ and $\lambda_2(\mathcal{Y}) = \sqrt{\lambda_1(\mathcal{H})}$ for $\mathcal{Y} = \begin{bmatrix} 0 & \mathcal{X} \\ \mathcal{X}^T & 0 \end{bmatrix}$, we get all inequalities, by Proposition A.6 and the following: $\sqrt{(\lambda_1(\mathcal{Y})^2 + \lambda_2(\mathcal{Y})^2)} \leq \lambda_1(\mathcal{Y}) + \lambda_2(\mathcal{Y}) = \sqrt{2(\lambda_1(\mathcal{Y})^2 + \lambda_2(\mathcal{Y})^2)}$. □

### A.3. Proofs of Proposition 2.5 and 2.6

*Proof of Proposition 2.5.* We rewrite Proposition A.5 given that, for a normalized $\mathcal{X}$, $\mathsf{gap}(\mathcal{H}) \geq \frac{8c_0 NK}{5}$, as Proposition A.8 and $\lambda_j(\mathcal{Y}) = s_j(\mathcal{X})$. In particular,

$$\begin{aligned}
\mathsf{gap}(\mathcal{X}) &= \mathsf{gap}(\mathcal{Y}) = \frac{\mathsf{gap}(\mathcal{H})}{\lambda_1(\mathcal{Y}) + \lambda_2(\mathcal{Y})} \\
&\geq \frac{\mathsf{gap}(\mathcal{H})}{\sqrt{N_1 a + N_2 c}} \geq \frac{8c_0 NK}{5\sqrt{2NK}} \\
&\geq \frac{4\sqrt{2NK}}{5}.
\end{aligned}$$

For the upper bound on $\mathsf{gap}(\mathcal{X})$, we have that

$$\begin{aligned}
\mathsf{gap}(\mathcal{X}) &= \mathsf{gap}(\mathcal{Y}) = \frac{\mathsf{gap}(\mathcal{H})}{\lambda_1(\mathcal{Y}) + \lambda_2(\mathcal{Y})} \\
&\leq \frac{N_1 a + N_2 c}{\sqrt{N_1 a + N_2 c}} \leq \sqrt{2NK}
\end{aligned}$$

□

**Definition A.7.** *For our application, we have $\forall k$, $1 \geq p_1^k, p_2^k \geq 0$, and*

$$\mathcal{X} = \begin{bmatrix} \frac{1+p_1^1}{2} & \frac{1+p_1^2}{2} & \cdots & \frac{1+p_1^K}{2} \\ \cdots & & & \\ \frac{1+p_1^1}{2} & \frac{1+p_1^2}{2} & \cdots & \frac{1+p_1^K}{2} \\ \frac{1+p_2^1}{2} & \frac{1+p_2^2}{2} & \cdots & \frac{1+p_2^K}{2} \\ \cdots & & & \\ \frac{1+p_2^1}{2} & \frac{1+p_2^2}{2} & \cdots & \frac{1+p_2^K}{2} \end{bmatrix}_{2N \times K}$$

It is easy to see that with this normalized random matrix, $\lambda_1(\mathcal{H}) = \lambda_2(\mathcal{H})$ is not possible, given that $a, b, c \in [K/4, K]$; furthermore, $\mathsf{gap}(\mathcal{H}) = \Theta(NK)$ as in the Proposition A.8.

**Proposition A.8.** *Given $\mathcal{H} = \mathcal{X}\mathcal{X}^T$ and $a, b, c$ as in (2.8) for any expected value mean matrix $\mathcal{X}$, which is not necessarily normalized,*

$$\mathsf{gap}(\mathcal{H}) = \sqrt{(N_1 a - N_2 c)^2 + 4N_1 N_2 b^2} \geq \frac{8c_0 NK}{5},$$

*where $c_0 = \frac{|b|\sqrt{ac}}{K(a+c)}$.*

*Hence for a normalized $\mathcal{X}$, $\mathsf{gap}(\mathcal{H}) = \Theta(NK)$ given that $a, b, c \in [K/4, K]$.*

*Proof.* For a tighter lower bound of $\mathsf{gap}(\mathcal{H})$ than the obvious $|N_1 a - N_2 c|$, let us assume w.l.o.g. that $N_2 c \geq N_1 a$. Thus we have

$$N_2 \geq 2N\frac{a}{a+c} \tag{A.19}$$

We differentiate two cases:

- Balanced case: $N_1 a \geq \frac{4}{25} N_2 c$.
- Imbalanced case: $N_1 a \leq \frac{4}{25} N_2 c$.

For balanced case: we have $N_1 \geq \frac{4}{25} \frac{N_2 c}{a}$ and hence

$$\begin{aligned}
\mathsf{gap}(\mathcal{H}) &\geq \sqrt{4 N_1 N_2 b^2} \geq \frac{4N_2 |b|}{5}\sqrt{\frac{c}{a}} \\
&\geq \frac{8N|b|}{5}\frac{a}{a+c}\sqrt{\frac{c}{a}} \geq \frac{8N|b|}{5}\frac{\sqrt{ac}}{a+c} \\
&= \frac{8c_0 NK}{5},
\end{aligned}$$

where $N_2 \geq 2N\frac{a}{a+c}$ as in (A.19).

For the imbalanced case, given that $\sqrt{ac} \geq |b|$ by Proposition 2.10,

$$\begin{aligned}
\mathsf{gap}(\mathcal{H}) &\geq \sqrt{(N_1 a - N_2 c)^2} \geq \frac{21}{25} N_2 c \\
&\geq \frac{42}{25}\frac{Nac}{a+c} \geq \frac{8}{5}\frac{N|b|\sqrt{ac}}{a+c} \geq \frac{8c_0 NK}{5}.
\end{aligned}$$

Finally, for a normalized random matrix $X$ and its $\mathcal{X}$, we have $c_0$ being a constant and combing with the upper bound of $\mathsf{gap}(\mathcal{H}) \leq N_1 a + N_2 c \leq 2NK$ concludes that $\mathsf{gap}(\mathcal{H}) = \Theta(NK)$. □

*Proof of Proposition 2.6.* By (A.2), $\bar{u}_1, \bar{u}_2$ are the first and second eigenvectors of $\mathcal{H}$ corresponding to $\lambda_1(\mathcal{H})$ and $\lambda_2(\mathcal{H})$. Let $x, y$ be entries that correspond to $P_1, P_2$ respectively in the first or second eigenvectors of $\mathcal{H}$. By (A.9) and (A.10), we have

$$\frac{y}{x} = \frac{\lambda - N_1 a}{N_2 b} = \frac{N_1 b}{\lambda - N_2 c}.$$

In addition, given any $b \neq 0$, we have $\mathsf{gap}(\mathcal{H}) > |N_1 a - N_2 c|$ and hence $\lambda_1(\mathcal{H}) > \max\{N_1 a, N_2 c\} > \lambda_2(A)$. Therefore, for $b > 0$, $\frac{y}{x} \geq 0$ for first eigenvector and $\leq 0$ for $v_2$. and for $b < 0$, it is the opposite. □

### *A.4. Proof of Lemma 2.7*

*Proof of Lemma 2.7.* We first show that $|x_2|, |y_2|$ are within a constant factor of each other, given that $\omega_1/\omega_2 = \frac{N_1}{N_2}$ is a constant.

**Proposition A.9.** *For a normalized $\mathcal{X}$, where $N_1, N_2, a, b \neq 0$, $x_2, y_2$ in the second top left singular vector $\bar{u}_2$ satisfy*

$$\frac{2N_2}{N_1} \geq \frac{|x_2|}{|y_2|} \geq \frac{N_2}{2N_1}$$

*Proof.* By (A.9) and given the upper bound on $\mathsf{gap}(\mathcal{H})$ in (A.8),

$$\frac{|y_2|}{|x_2|} = \frac{N_1 a - \lambda_2}{N_2 b} = \frac{N_1 a - N_2 c + \mathsf{gap}(\mathcal{H})}{2N_2 b} \leq \frac{N_1 a}{N_2 b}, \tag{A.20}$$

and hence $\frac{|x_2|}{|y_2|} \geq \frac{N_2 b}{N_1 a}$. By (A.10) and (A.8), we have

$$\frac{|x_2|}{|y_2|} = \frac{N_2 c - \lambda_2}{N_1 b} = \frac{N_2 c - N_1 a + \mathsf{gap}(\mathcal{H})}{2N_1 b} \leq \frac{N_2 c}{N_1 b} \tag{A.21}$$

We finish the proof by observing that $\frac{1}{2} \leq \frac{a}{b} \leq 2$, $\frac{1}{2} \leq \frac{c}{b} \leq 2$:, due to the fact that $\frac{1}{2} \leq \frac{\mu_i^j}{\mu_2^j} \leq 2, \forall j = 1, \ldots, K$ for $\mu_i^j \in [1/2, 1]$ in a normalized $\mathcal{X}$, and the following lemma:

**Lemma A.10.** *If $0 < c_{\min} \leq \frac{a_i}{b_i} \leq c_{\max}, \forall i = 1, \ldots, n$, where $a_i, b_i > 0$, then $c_{\min} \leq \frac{\sum_{i=1}^n a_i}{\sum_{i=1}^n b_i} \leq c_{\max}$.*

*Proof.* $c_{\min} = \frac{\sum_{i=1}^n c_{\min} b_i}{\sum_{i=1}^n b_i} \leq \frac{\sum_{i=1}^n a_i}{\sum_{i=1}^n b_i} \leq \frac{\sum_{i=1}^n c_{\max} b_i}{\sum_{i=1}^n b_i} = c_{\max}.$ □

□

Let $x = x_2$ and $y = y_2$. By Proposition A.9, $|y| \leq \frac{2N_1|x|}{N_2}$ and

$$1 = N_1 x^2 + N_2 y^2 \leq N_1 x^2 + N_2 \left(\frac{2|x|N_1}{N_2}\right)^2 \leq x^2 \left(\frac{4N_1^2 + N_1 N_2}{N_2}\right),$$

hence for $C_{x\,\min} = \frac{\omega_2}{4\omega_1^2 + \omega_1 \omega_2}$,

$$x^2 \geq \frac{\omega_2}{4\omega_1^2 + \omega_1 \omega_2} \frac{1}{2N}. \tag{A.22}$$

Looking in the other direction, by Proposition A.9, $|x| \leq \frac{2|y|N_2}{N_1}$,

$$1 = N_1 x^2 + N_2 y^2 \leq N_2 y^2 + N_1 \left(\frac{2|y|N_2}{N_1}\right)^2 \leq y^2 \left(\frac{N_2 N_1 + 4N_2^2}{N_1}\right),$$

and hence for a given $C_{y\min} = \frac{\omega_1}{4\omega_2^2+\omega_1\omega_2}$,

$$|y|^2 \geq \frac{\omega_1}{4\omega_2^2 + \omega_1\omega_2} \frac{1}{2N}.$$

On the other hand, by Proposition A.9, we have $|y| \geq \frac{N_1|x|}{2N_2}$, we have

$$1 = N_1 x^2 + N_2 y^2 \geq N_1 x^2 + N_2 \left(\frac{|x|N_1}{2N_2}\right)^2 \geq x^2 \left(\frac{N_1^2 + 4N_1 N_2}{4N_2}\right),$$

and thus $x^2 \leq \frac{4\omega_2}{\omega_1^2+4\omega_1\omega_2}\frac{1}{2N}$. Looking in the other direction, by Proposition A.9, $|x| \geq \frac{|y|N_2}{2N_1}$,

$$1 = N_1 x^2 + N_2 y^2 \geq N_2 y^2 + N_1 \left(\frac{|y|N_2}{2N_1}\right)^2 \geq y^2 \left(\frac{4N_2 N_1 + N_2^2}{4N_1}\right),$$

and hence $|y|^2 \leq \frac{4\omega_1}{\omega_2^2+4\omega_1\omega_2}\frac{1}{2N}$. Hence we have that

$$|x-y|^2 = (|x|+|y|)^2 \leq \left(\sqrt{\frac{4\omega_2}{\omega_1^2+4\omega_1\omega_2}\frac{1}{2N}} + \sqrt{\frac{4\omega_1}{\omega_2^2+4\omega_1\omega_2}\frac{1}{2N}}\right)^2$$

$$\leq \frac{1}{2N}\left(\sqrt{\frac{4\omega_2}{\omega_1^2+4\omega_1\omega_2}} + \sqrt{\frac{4\omega_1}{\omega_2^2+4\omega_1\omega_2}}\right)^2,$$

and $C_{\max} = (\sqrt{\frac{1}{\omega_1}} + \sqrt{\frac{1}{\omega_2}})^2$. Hence

$$C_{\max} \leq \left(\sqrt{\frac{4\omega_2}{\omega_1^2+4\omega_1\omega_2}} + \sqrt{\frac{4\omega_1}{\omega_2^2+4\omega_1\omega_2}}\right)^2 \leq \left(\sqrt{\frac{1}{\omega_1}} + \sqrt{\frac{1}{\omega_2}}\right)^2.$$

□

### Appendix B: Proof of Lemma 2.16

Recall that the largest left singular vectors $u_1, u_2$ has the form of $[x, \ldots, x, y, \ldots, y]$, where $x$ repeats $N_1$ times and $y$ repeats $N_2$ times. *Proof of Lemma 2.16.* Let us define the following random variables,

$$\delta = \frac{1}{N_1}\sum_{i=1}^{N_1}\delta_i, \qquad \tau = \frac{1}{N_2}\sum_{i=1}^{N_2}\tau_i, \qquad (B.1)$$

such that by Claim 2.15,

$$|\delta| = \left|\frac{1}{N_1}\sum_{i=1}^{N_1}\delta_i\right| \leq \frac{1}{N_1}\sum_{i=1}^{N_1}|\delta_i| \leq \frac{\sqrt{N_1\sum_{i=1}^{N_1}\delta_i^2}}{N_1} \leq \frac{c_1\sigma}{\sqrt{N_1 N}}$$

$$|\tau| = \left|\frac{1}{N_2}\sum_{i=1}^{N_2}\tau_i\right| \leq \frac{1}{N_2}\sum_{i=1}^{N_2}|\tau_i| \leq \frac{\sqrt{N_2\sum_{i=1}^{N_2}\tau_i^2}}{N_2} \leq \frac{c_1\sigma}{\sqrt{N_2 N}}$$

and hence

$$\max(|N_1\delta|, |N_2\tau|) \leq \frac{c_1\sigma\sqrt{N_2}}{\sqrt{N}} \qquad (B.2)$$

given that we always assume that $N_2 > N_1$. A natural classifier to separate individuals would be: $\frac{x+y}{2}$ when we use $u_1$; but we do not have access to $x$ and $y$. Recall that

$$M = \frac{\sum_{i=1}^{N_1}(x+\delta_i) + \sum_{i=1}^{N_2}(y+\tau_i)}{2N} = \frac{N_1 x + N_2 y}{2N} + \frac{N_1\delta + N_2\tau}{2N}.$$

We are now ready to show that when $N_1, N_2$ are large enough, we see enough separation between the mixture sample mean and both $x$ and $y$. We first prove the following claims.

**Claim B.1.** $xN_1\delta + yN_2\tau = -\frac{\|\epsilon\|_2^2}{2}$.

*Proof.* This claim is obvious given that $\|u_1\|_2 = \|\bar{u}_1\|_2 = 1$, and $\bar{u}_1, u_1, \epsilon$ all being real vectors, $\|u_1\|_2^2 = \|\bar{u}_1\|_2^2 + \|\epsilon\|_2^2 + 2 <\bar{u}_1, \epsilon> = \|\bar{u}_1\|_2^2 + \|\epsilon\|_2^2 + 2xN_1\delta + yN_2\tau$. □

We next use $\frac{1}{\sqrt{2N}}|xN_1\delta + yN_2\tau|$ to obtain a bound on $\left|\frac{N_1\delta + N_2\tau}{2N}\right|$, given that

$$\frac{1}{\sqrt{2N}}|xN_1\delta + yN_2\tau| \leq \frac{\|\epsilon\|_2^2}{2\sqrt{2N}} \leq \frac{c_1^2\sigma^2}{2N\sqrt{2N}}; \qquad (B.3)$$

**Claim B.2.** Let $N_1 \leq N_2$, and $\omega_1 = \frac{N_1}{2N}$ and $\omega_2 = \frac{N_2}{2N}$, and given that $N_1 \geq \max\left(\frac{2c_1^2\sigma^2}{c_2\gamma}, \frac{8c_1^2\sigma^2}{\gamma}\frac{\omega_2}{\omega_1}\right)$, we have

$$\left|\frac{N_1\delta + N_2\tau}{2N}\right| \leq \frac{N_1|y-x|\sqrt{\gamma}}{2N} \qquad (B.4)$$

*Proof.* We next derive a bound on $\frac{N_1\delta + N_2\tau}{2N}$. By Separation Lemma 2.8, we have $|x-y| = c_2\sqrt{\frac{\gamma}{2N}}$ for a constant $c_2 = 1/2$, and thus we have $\frac{\max(x,y)}{\sqrt{2N}} > \frac{1}{2N}$. Therefore,

$$\frac{|xN_1\delta + yN_2\tau|}{\sqrt{2N}}$$
$$= \frac{|\max(x,y)(N_1\delta + N_2\tau) + (x - \max(x,y))N_1\delta + (y - \max(x,y))N_2\tau|}{\sqrt{2N}}$$
$$\geq \frac{|\max(x,y)|(N_1\delta + N_2\tau)}{\sqrt{2N}} - \frac{|x-y|\max(|N_1\delta|, |N_2\tau|)}{\sqrt{2N}}$$

Thus we have, given (B.2), (B.3) and (B.5),

$$
\begin{aligned}
\left| \frac{N_1\delta + N_2\tau}{2N} \right| &\leq \frac{|\max(x,y)|}{\sqrt{2N}} |(N_1\delta + N_2\tau)| \\
&\leq \frac{|xN_1\delta + yN_2\tau|}{\sqrt{2N}} + \frac{|x-y|\max(|N_1\delta|,|N_2\tau|)}{\sqrt{2N}} \\
&\leq \frac{c_1^2 \sigma^2}{2N\sqrt{2N}} + \frac{c_2\sqrt{\gamma}}{2N} c_1\sigma \sqrt{\frac{N_2}{N}} \\
&\leq \frac{N_1 c_2 \gamma}{2N\sqrt{2N}} \leq \frac{N_1|y-x|\sqrt{\gamma}}{2N},
\end{aligned}
$$

where

$$
\frac{c_1^2 \sigma^2}{2N\sqrt{2N}} < \frac{N_1 c_2 \gamma}{4N\sqrt{2N}}, \text{ holds so long as } N_1 \geq \frac{2c_1^2 \sigma^2}{c_2 \gamma}, \text{ and} \quad \text{(B.5)}
$$

$$
\frac{c_2\sqrt{\gamma} c_1 \sigma}{2N} \sqrt{\frac{N_2}{N}} < \frac{N_1 c_2 \gamma}{4N\sqrt{2N}}. \text{ holds so long as } N_1 \geq \frac{8 c_1^2 \sigma^2}{\gamma} \frac{\omega_2}{\omega_1}, \text{ so that} \quad \text{(B.6)}
$$

$$
N_1 \geq \frac{2\sqrt{2} c_1 \sigma \sqrt{N_2}}{\sqrt{\gamma}}, \quad \text{(B.7)}
$$

Both conditions are guaranteed by (2.23) in Theorem 2.14. □

This allows us to conclude that

$$
\begin{aligned}
\left| \frac{N_1 x + N_1\delta + N_2 y + N_2\tau}{2N} - \frac{N_1 x + N_2 y}{2N} \right| &= \left| \frac{N_1\delta + N_2\tau}{2N} \right| \\
&\leq \left( \frac{\min\{N_1, N_2\}\sqrt{\gamma}}{2N} \right) |x-y|.
\end{aligned}
$$

Given that $|y-x| = c_2\sqrt{\gamma}/\sqrt{2N}$ as shown in the Separation Lemma 2.8, we have

$$
\begin{aligned}
\left| \frac{N_1 x + N_1\delta + N_2 y + N_2\tau}{2N} - x \right| &= \left| \frac{N_2(y-x)}{2N} + \frac{N_1\delta + N_2\tau}{2N} \right| \\
&\geq \left| \left| \frac{N_2(y-x)}{2N} \right| - \left| \frac{N_1\delta + N_2\tau}{2N} \right| \right| \\
&\geq \frac{N_2|y-x|}{2N} - \frac{\min\{N_1, N_2\}\sqrt{\gamma}|y-x|}{2N} \\
&\geq \frac{(1-\sqrt{\gamma})N_2|y-x|}{2N},
\end{aligned}
$$

and similarly,

$$\left|y - \frac{N_1 x + N_1 \delta + N_2 y + N_2 \tau}{2N}\right| = \left|\frac{N_1(y-x)}{2N} - \frac{N_1 \delta + N_2 \tau}{2N}\right|$$
$$\geq \left|\frac{N_1(y-x)}{2N}\right| - \left|\frac{N_1 \delta + N_2 \tau}{2N}\right|$$
$$\geq \frac{N_1|y-x|}{2N} - \frac{\min\{N_1, N_2\}\sqrt{\gamma}|y-x|}{2N}$$
$$\geq \frac{(1-\sqrt{\gamma})N_1|y-x|}{2N}.$$

□